\documentclass[10pt,twocolumn,letterpaper]{article}

\usepackage{cvpr}      
\usepackage{graphicx}
\usepackage{amsmath}
\usepackage{amssymb}
\usepackage{booktabs}
\usepackage{lipsum}
\usepackage{adjustbox}
\usepackage{subcaption}

\usepackage{mathrsfs}
\usepackage{makecell}

\usepackage{footmisc}

\usepackage{array}
\usepackage{times}
\usepackage{epsfig}
\usepackage{graphicx}
\usepackage{float}
\usepackage{wrapfig}
\usepackage{amsmath,amssymb,amsthm}
\usepackage{algorithm,algorithmicx,algpseudocode}
\usepackage{bm,xspace}
\usepackage{comment}
\usepackage{multirow}
\usepackage{balance}
\usepackage{url}
\usepackage{booktabs}
\usepackage{etoolbox,siunitx}
\usepackage{calc}
\usepackage{pifont,hologo}
\usepackage{color}
\usepackage{adjustbox}
\usepackage[normalem]{ulem}  
\usepackage[table]{xcolor}

\definecolor{lightgreen}{HTML}{D8ECD1}
\definecolor{darkgreen}{rgb}{0.13, 0.55, 0.13}

\newcommand{\authorskip}{\hspace{4mm}}

\definecolor{blue}{HTML}{0055cc}
\definecolor{red}{HTML}{cc1100}
\definecolor{orange}{HTML}{cc7700}
\definecolor{gray}{HTML}{efefef}
\definecolor{darkgreen}{rgb}{0.13, 0.55, 0.13}
\definecolor{darkgray}{HTML}{757575}

\renewcommand{\eqref}[1]{Eq.~\ref{#1}}

\newcolumntype{x}[1]{>{\centering\arraybackslash}p{#1}}
\newcolumntype{y}[1]{>{\raggedright\arraybackslash}p{#1}}
\newcolumntype{z}[1]{>{\raggedleft\arraybackslash}p{#1}}

\DeclareMathSymbol{@}{\mathord}{letters}{"3B}

\newcommand\mypara[1]{\vspace{0mm}\noindent\textbf{#1}}

\makeatletter
\DeclareRobustCommand\onedot{\futurelet\@let@token\@onedot}
\def\@onedot{\ifx\@let@token.\else.\null\fi\xspace}

\newcommand*{\Rom}[1]{\expandafter\@slowromancap\romannumeral #1@}
\newcommand*{\rom}[1]{\expandafter\romannumeral #1}

\def\1{\bm{1}}

\def\mP{{\bm{P}}}

\def\mT{{\bm{T}}}

\DeclareMathAlphabet{\mathsfit}{\encodingdefault}{\sfdefault}{m}{sl}
\SetMathAlphabet{\mathsfit}{bold}{\encodingdefault}{\sfdefault}{bx}{n}

\usepackage{tcolorbox}
\newtcolorbox{myboxnote}[1][]{
  title=#1,
  colback=blue!2,
  colbacktitle=blue!2,
  coltitle=black,
  fonttitle=\bfseries,
  bottomrule=0pt,
  toprule=0pt,
  leftrule=1.5pt,
  rightrule=1.5pt,
  titlerule=0pt,
  arc=0pt,
  outer arc=0pt,
  colframe=blue!30,
}

\definecolor{cvprblue}{rgb}{0.21,0.49,0.74}
\usepackage[pagebackref,breaklinks, colorlinks, citecolor=cvprblue]{hyperref}
\usepackage[hypcap=true]{caption}

\begin{document}


\title{\vspace{-5mm}GPT4Point: A Unified Framework for Point-Language \\ Understanding and Generation}
\vspace{-8mm}
\author{
Zhangyang Qi\textsuperscript{1}\footnotemark[1] \authorskip 
Ye Fang\textsuperscript{2,5}\footnotemark[1] \authorskip 
Zeyi Sun\textsuperscript{3,5}\footnotemark[1] \authorskip \\
Xiaoyang Wu\textsuperscript{1} \authorskip 
Tong Wu\textsuperscript{4} \authorskip 
Jiaqi Wang\textsuperscript{5}\footnotemark[2] \authorskip 
Dahua Lin\textsuperscript{4,5} \authorskip 
Hengshuang Zhao\textsuperscript{1}\footnotemark[2] \\
\normalsize{$^*$ Equal contribution}\quad  $\dagger$ Corresponding author\vspace{0.5mm}\\
\textsuperscript{1}The University of Hong Kong \authorskip 
\textsuperscript{2}Fudan University \authorskip 
\textsuperscript{3}Shanghai Jiao Tong University \authorskip \\
\textsuperscript{4}The Chinese University of Hong Kong
\textsuperscript{5}Shanghai AI Laboratory \\
{\tt\small \{zyqi, xywu3, hszhao\}@cs.hku.hk, wangjiaqi@pjlab.org.cn} \\
\small \url{https://gpt4point.github.io}
}


\twocolumn[{%
\renewcommand\twocolumn[1][]{#1}%
\maketitle
\begin{center}
\centering
\vspace{-10mm}
\includegraphics[width=0.98\textwidth]{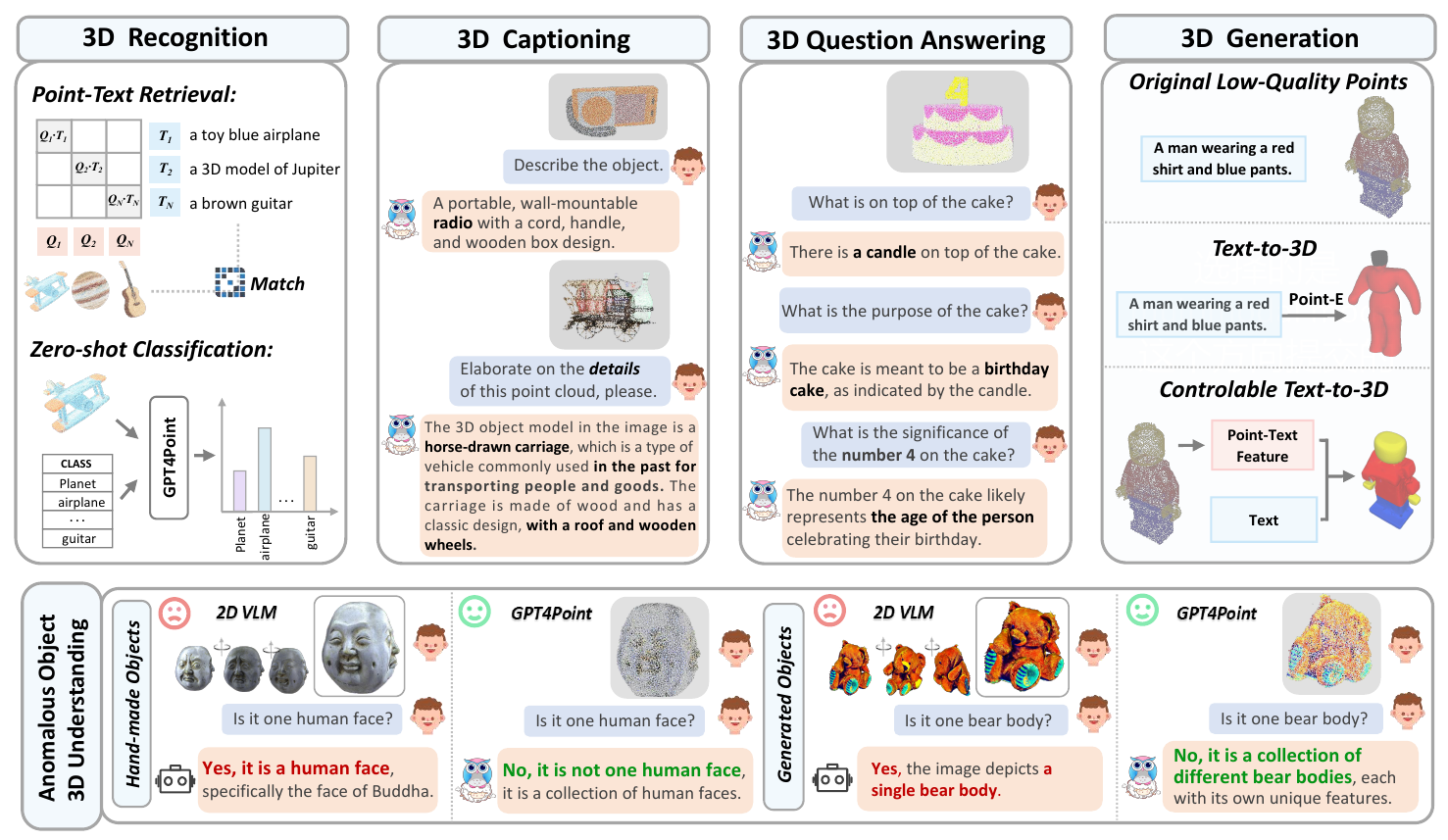}
\label{fig_1:teaser}
\vspace{-3mm}
\captionof{figure}
{
\textbf{Task examples of GPT4Point}. It performs accurate 3D recognition, detailed captioning, precise Q\&A, and high-quality controllable 3D generation. Additionally, GPT4Point excels in 3D anomalous object description, accurately assessing abnormal shapes like the multi-face object and the 3D generation failure case. It is a crucial ability in the assessment of generated 3D objects. 
}
\label{fig1_teaser}
\end{center}
\vspace{-2mm}
}]

\begin{abstract}
\vspace{-4mm}
Multimodal Large Language Models (MLLMs) have excelled in 2D image-text comprehension and image generation, but their understanding of the 3D world is notably deficient, limiting progress in 3D language understanding and generation. To solve this problem, we introduce GPT4Point, an innovative groundbreaking point-language multimodal model designed specifically for unified 3D object understanding and generation within the MLLM framework. GPT4Point as a powerful 3D MLLM seamlessly can execute a variety of point-text reference tasks such as point-cloud captioning and Q\&A. Additionally, GPT4Point is equipped with advanced capabilities for controllable 3D generation, it can get high-quality results through a low-quality point-text feature maintaining the geometric shapes and colors. To support the expansive needs of 3D object-text pairs, we develop Pyramid-XL, a point-language dataset annotation engine. It constructs a large-scale database over 1M objects of varied text granularity levels from the Objaverse-XL dataset, essential for training GPT4Point. A comprehensive benchmark has been proposed to evaluate 3D point-language understanding capabilities. In extensive evaluations, GPT4Point has demonstrated superior performance in understanding and generation.
\end{abstract}
\vspace{-10pt}

\newpage
 
\section{Introduction}

~~~The recent Large Language Models (LLMs)~\cite{ChatGPT, GPT4, InstructGPT, llama, vicuna, internlm, flan, t5} have demonstrated remarkable advancements in the field of natural language processing. Inspired by their powerful capabilities, researchers have also explored Multimodal LLMs (MLLMs), via adapting LLMs into various modalities like images~\cite{BLIP-2, LLaVA}, audio~\cite{Audiogpt, TANGO, Llm4ts} and videos~\cite{VideoBLIP, VideoLLM}. The proliferation of extensive image-text pair~\cite{cc12m, laion} has crucially enabled 2D MLLMs \textit{i.e.}, Vision Language Models (VLMs) to interpret images through textual representations. Concurrently, there is a growing trend in utilizing these multimodal models for guiding text-to-image generation~\cite{SEED, Emu, MT, CM3Leon}. This represents a form of compression and reconstruction, exploring how to accurately recover and even edit the input image using controllable image generation models. However, despite the impressive capabilities of MLLMs in handling multiple modalities, they still face significant limitations in understanding and accurately interpreting the 3D world, a critical need for various important downstream applications like intelligent robotics and augmented reality.

\vspace{2pt}
Recent efforts to develop 3D MLLMs~\cite{3d-llm, 3d-vista} have notable limitations. Some ~\cite{3d-llm, 3d-vista} prioritize the overall scene and focus primarily on the spatial coordinates of objects, often neglecting the geometric details of individual objects. This can lead to a limited understanding of the 3D world. Meanwhile, these methods generally convert 2D image features into 3D representations~\cite{3d-llm}, which leads to a substantial loss of geometric accuracy. 3D geometry information is important in understanding. As shown at the bottom of~\cref{fig1_teaser}, the VLM fails to recognize the four-sided face object while our GPT4Point can figure out the anomalies. Concurrent works focusing on utilizing 3D features directly exhibit notable limitations. PointBind~\cite{Pointbind} exhibits a deficiency in training and demonstrates restricted text referencing abilities due to the limited dataset. On the other hand, PointLLM~\cite{Pointllm} necessitates the training of the corresponding Language Model (LLM) component and does not possess the capability to expand into text generation.

\vspace{2pt}
We present GPT4Point\footnote{First author is the intern at Shanghai AI Laboratory.}, a novel unified framework for point-language understanding and generation. GPT4Point introduces the 3D object MLLM, which is a groundbreaking language model that fully utilizes point clouds to perform various point-text tasks as shown in~\cref{fig1_teaser}. We utilize a Bert-based Point-QFormer for point-text feature alignment. Aligned features are separately input into the LLMs for text inference tasks and Diffusion for 3D object generation tasks. It is worth noting that, given a low-quality point cloud feature as a condition, GPT4Point can generate higher-quality results while maintaining the geometric shapes and colors by using point-text aligned features called controllable text-to-3D.

\vspace{3pt}
To tackle the scarcity of object point-language data~\cite{OmniObject3D}, we leverage the Objaverse-XL dataset~\cite{Objaverse, Objaverse-XL} to develop an automated, effective data annotation engine Pyramid-XL. It employs Vision Language Models (VLMs) for generating text annotations. Pyramid-XL solves the problem that VLMs can not understand multi-view images directly. By synthesizing captions from multi-views obtained by the VLMs, the text annotation is stratified into three hierarchical levels, ranging from low to high, ultimately leading to precise annotations. Apart from the data engine, we establish an object point-text benchmark for assessing point multimodal model capabilities in recognition and text inference tasks, such as 3D object point cloud captioning, and Q\&A. This benchmark also provides a critical standard for evaluating 3D object generation, while current assessments often rely on qualitative judgments from rendered images without a direct evaluation in 3D space~\cite{DreamFusion}. Only relying on rendering images may lead to misunderstanding, for instance, in the bottom right of~\cref{fig1_teaser},  a failure case produced by 3D generation (a bear has two bodies), makes 2D VLMs and even humans fail to recognize its anomaly but our model can identify with anomalies easily.

\vspace{3pt}
\noindent Our paper makes three major contributions:
\begin{itemize}
\item We present the unified framework for point-language understanding and generation \textbf{GPT4Point}, including the 3D MLLM for point-text tasks and controlled 3D generation.
\item Introducing the automated point-language dataset annotation engine \textbf{Pyramid-XL} based on Objaverse-XL, currently encompassing 1M pairs of varying levels of coarseness and can be extended cost-effectively.
\item Establishing a novel object-level point cloud benchmark with comprehensive evaluation metrics for 3D point cloud language tasks. This benchmark thoroughly assesses models' understanding capabilities and facilitates the evaluation of generated 3D objects.
\end{itemize}

\begin{figure*}[!t]
    \centering
    \includegraphics[width=\linewidth]{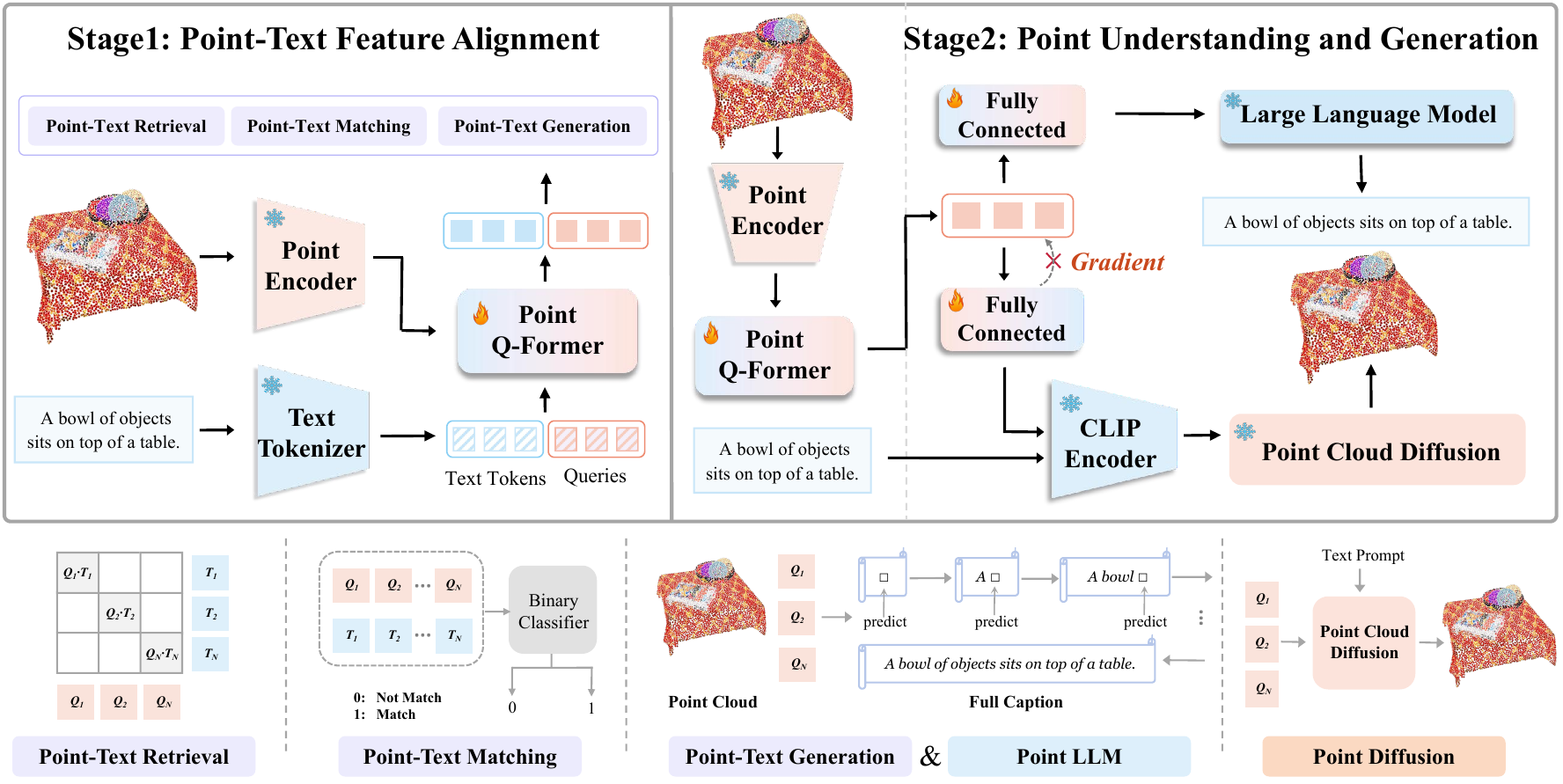}
    \vspace{0.2mm}
    \caption{\textbf{The model architecture of GPT4Point for training.} In Stage1, we employ a Bert-based~\cite{Bert} Point-Q-Former for point-text feature alignment through three point-text tasks. Then, in Stage2, an LLM is appended to train the model's text inference capabilities. A Point Cloud Diffusion is attached separately to train controlled text-to-3D generation which keeps the geometry shape and colors.}
    \label{fig2_architechure}
    \vspace{-2mm}
\end{figure*}

\section{Related Work}
\noindent \textbf{Multi-modal large language models (MLLMs).} Large Language Models (LLMs) have demonstrated robust capabilities in language comprehension, reasoning, and generalization~\cite{ChatGPT, GPT4, InstructGPT, llama, vicuna, internlm, flan, t5}. Building upon this, Multimodal Large Language Models (MLLMs) extend these reasoning skills to additional modalities such as image~\cite{Palm-E, LLaMA-Adapter, LLaMA-Adapterv2, image-bind, meta-transformer, MiniGPT-4}, audio~\cite{Audiogpt, TANGO, Llm4ts}, and video~\cite{VideoChat, VideoBLIP, VideoLLM}. Typically, MLLMs align target features with textual features and then integrate them with LLMs for various text inference tasks. Some train the whole architecture from scratch~\cite{Kosmos-1, Kosmos-2} and others~\cite{Otter, Qwen-VL, LLaVA, BLIP-2, InstructBLIP} utilize pretrained LLMs. In the realm of 3D MLLMs, existing models either rely on 2D image information~\cite{3d-llm, 3d-vista} or simply align low-quality textual phrases with points~\cite{Point-bind, Pointllm}. To solve these problems, we introduce a novel 3D MLLM designed for diverse point-text tasks. Our model, featuring a Point Q-Former based on Bert~\cite{Bert}, aligns two domain features and integrates an LLM for text-based reasoning tasks, advancing the field of 3D multimodal understanding.

\vspace{4pt}
\noindent \textbf{Language-driven 3D object understanding.}
3D point cloud multimodal models encompass a broad spectrum, generally categorized into those focusing on the entire scene containing multiple objects and those focusing on individual objects. The former places more emphasis on the relative positions of objects in the scene rather than their geometric shapes; Here, we primarily focus on the latter. In a self-supervised way, powerful backbones like PointBert~\cite{Pointbert} for object points have been obtained~\cite{Pointbert, PointMAE}. Then, point cloud language pretraining attempts to align the point cloud modality and the text modality. Some methods~\cite{PointCLIP, CLIP2Point} try to convert point clouds to depth images for alignment with text using CLIP~\cite{CLIP}. Tri-modal approaches such as ULIP~\cite{ulip, ulip2, Pointbind, meta-transformer} integrate point cloud, text, and image data. However, these methods all exclusively use 2D images, either explicitly or implicitly. Our work differs by directly aligning 3D point-text modalities, completely removing the dependency on image data.

\vspace{4pt}
\noindent \textbf{Text-to-3D generation.} 
Text-to-image generation models have experienced significant advancements recently~\cite{Stable_Diffusion, ControlNet}, yet text-to-3D models face challenges due to limited 3D data availability. Current approaches often rely on optimizing Neural Radiance Fields (NeRF) representation~\cite{NeRF} with Score-Distillation-Sampling (SDS) loss~\cite{DreamFusion}. While these optimization-based methods~\cite{DreamFusion, Magic3D, Fantasia3d, ProlificDreamer} still fall short in robustness, speed, and generalization. Alternatively, Point-E~\cite{Point-E} and Shap-E~\cite{Shape-E} employ feed-forward 3D generative models trained on large, undisclosed 3D datasets, offering better generalization and faster processing. However, these models often produce random, uncontrollable outputs with low-quality textures. To solve these limitations, we leverage point-text features to enhance the controllability of feed-forward models. This approach uses a low-quality point-text feature as a condition that allows for maintaining specific shapes and colors, thereby enabling the generation of higher-quality 3D objects.

\begin{figure*}[!t]
    \centering
    \includegraphics[width=0.99\linewidth]{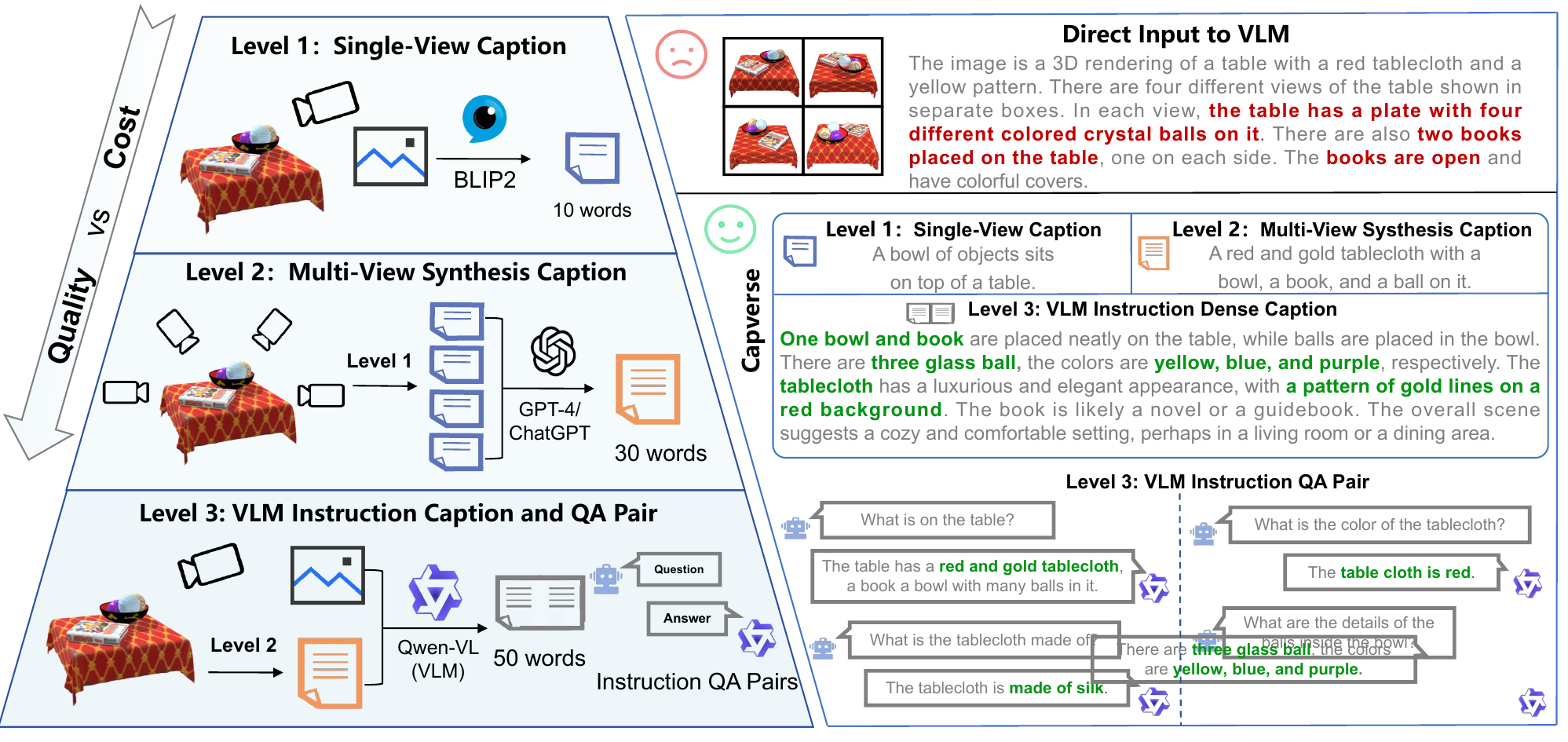}
    \vspace{-2mm}
    \caption{\textbf{Pyramid-XL: An automated point-text annotation engine.} Directly inputting images into VLMs yields unsatisfactory results. We propose a progressive annotation approach with 3 levels of granularity, leveraging results from the previous level for precise outcomes. }
    \label{fig3_dataset_engine}
    \vspace{-2mm}
\end{figure*}

\section{Methods}
~~~~This section provides an overview of our data text annotation engine and model architecture. In~\cref{sec_3.1:Point-Language Annotation Engine}, we introduce \textbf{Pyramid-XL}, our point-language dataset annotation engine, discussing its design, function, and the progression from low-quality descriptions to ultimately precise and detailed ones. Then, in~\cref{sec_3.2:Model Architecture}, we delve into GPT4Point's architecture, explaining how to align point and text and demonstrating how LLM and point diffusion models contribute to unified understanding and generation.

\newpage
\subsection{Point-Language Dataset Annotation Engine}
\label{sec_3.1:Point-Language Annotation Engine}

~~The public release of the large-scale Objaverse dataset~\cite{Objaverse} and its successor Objaverse-XL~\cite{Objaverse-XL} includes 800K and 10M objects respectively, providing a vast amount of 3D object data. However, these objects lack corresponding text descriptions. We plan to use the rendered images of the objects as input and obtain textual descriptions through a trained Vision Language Model (VLM), however, we find that direct input of multi-view images into the VLM does not enable it to understand their 3D structure and give precise descriptions, as shown in the top right of~\cref{fig3_dataset_engine}. Hence, Pyramid-XL employs a hierarchical pipeline, evolving from initial low-quality descriptions to achieve ultimately precise and detailed results.

\begin{myboxnote}[Pyramid-XL]
\textbf{\textit{Single-View Caption (Level 1)}}: We use the primary VLM model BLIP-2~\cite{BLIP-2} to generate concise descriptions, approximately 10 words in length, from a single-view rendered image.

\textbf{\textit{Multi-View Caption (Level 2)}}: This level synthesizes multiple Level 1 descriptions by GPT-4~\cite{GPT4} to create comprehensive multi-view captions which has approximately 30 words.

\textbf{\textit{VLM Instruction Caption and QA Pair (Level 3)}}: Utilizing the view with the highest CLIP score, selected from textual descriptions, we engage the advanced VLM to produce detailed dense captions and a corresponding QA dataset.
\end{myboxnote}

\vspace{6pt}
In terms of scale, Pyramid-XL is employed to annotate over 1M objects with Level 1 captions, 660K objects with Level 2 captions (same as Cap3D~\cite{Cap3D}), and 70K objects with Dense Captions including QA data. To assess the impact of text granularity on training, we designate the 1M Level 1 captions as the pretrain dataset, while a smaller set of detailed Level 3 data is used for instruction tuning. This methodology mirrors practices in the vision field, where models are initially pretrained on large volumes of coarser data and subsequently finetuned on more detailed data from specialized domains. Detailed experimental results of this approach are presented in Sec.~\ref{sec_5.3_Assessing the Effectiveness of Pyramid-XL}.

\subsection{Model Architecture}
\label{sec_3.2:Model Architecture}
~~~~GPT4Point consists of two stages as illustrated in~\cref{fig2_architechure}. In Stage1, we focus on point-text alignment using the Point-QFormer, a Bert-based structure similar to the Q-Former in BLIP-2~\cite{BLIP-2}. This stage involves supervision through three tasks related to recognition and text reasoning. In Stage2, only the point cloud is input into the point encoder and Point-QFormer to obtain aligned features, which are then devided into two branches: the LLM Branch and the Diffusion Branch separately. These branches supervise text comprehension and object generation tasks, respectively.

\vspace{6pt}
\noindent \textbf{Stage1: point-text feature alignment.}
Given a point cloud $\mP \in \mathbb{R}^{N \times 6}$, where each point is represented by six dimensions (XYZ coordinates and RGB color values), the initial stage of training focuses on feature extraction. The point encoder $\mathcal{E}$ processes the point cloud to yield the point cloud feature token $\mT_{1}^{p} = \mathcal{E}(\mP)$. Concurrently, the input text goes through tokenization via the Point Q-Former's text tokenizer, resulting in the text feature token $\mT_{1}^{t}$. These tokens, $\mT_{1}^{p}$ and $\mT_{1}^{t}$, are then utilized as inputs for the Point Q-Former $\mathcal{F}_{Q}$, facilitating the fusion of point cloud and textual data. We jointly optimize three training objectives: Point-Text Contrast (PTC) and Point-Text Matching (PTM), both recognition tasks, along with Point Caption Generation (PTG), a text inference task designed for aligning point clouds with textual data. The formulas are as follows:
\begin{equation}
\label{eqn:l1}
    \mathcal{L}_{1} = \mathcal{F}_{Q}\left(\mT_{1}^{p},  \mT_{1}^{t} \right) = \mathcal{F}_{Q} \left( \mathcal{E}\left( \mP \right),  \mT_{1}^{t} \right)
\end{equation}
Here, $\mathcal{L}_{1}$ represents the loss for three tasks, and we have set the weight ratios between them all to 1. In the final layer of $\mathcal{E}$, a fully connected layer maintains consistency between the dimensions of $\mT_{1}^{p}$ and $\mT_{1}^{t}$. 

\vspace{5pt}
\mypara{Stage2: point understing and generation.}
After the point-text feature alignment, we proceed with understanding and generation tasks. It's important to note that here we only input the point cloud into the Point Encoder and Point Q-Former to obtain the aligned feature. For the understanding task, a Large Language Model (LLM) is integrated with the Point Q-Former. The semantically integrated point cloud features are represented as $\mT_{2}^{P} = \mathcal{F}_{Q}\left(\mT_{1}^{p}\right) = \mathcal{F}_{Q}\left(\mathcal{E}\left(\mP\right)\right)$. The textual feature tokens $\mT_{2}^{t}$ are obtained from the LLM's own tokenizer. The objective function is defined as follows:
\begin{equation}
\label{eqn:l2}
    \mathcal{L}_{2} = \mathcal{F}_{\textit{LLM}}\left(\mT_{2}^{p},  \mT_{2}^{t}\right) = \mathcal{F}_{\textit{LLM}}\left(\mathcal{F}_{Q}\left(\mathcal{E}\left(\mP \right) \right),  \mT_{2}^{t} \right)
\end{equation}
$\mathcal{F}_{Q}$ indicates Point Q-former including a fully connected layer in its last layer to ensure consistency between the dimensions of $\mT_{2}^{p}$ and $\mT_{2}^{t}$. $\mathcal{L}_2$ represents the loss function from the Point Caption task alone. 

\vspace{5pt}
For 3D object generation, we utilize the features obtained from low-quality point clouds via the Point Q-Former as conditions inputted into the text-to-3D framework. This process results in the generation of refined 3D objects that maintain consistency in shape and color with the original point cloud. A notable distinction from the LLM branch is that we have not only frozen point cloud diffusion but also frozen Point Q-Former. As shown in \cref{fig2_architechure}, we employ a single fully-connected layer to project the aligned features into the CLIP token embedding space, referred to as $\mT_{3}^{p}$, and then concatenate these with the original text embeddings $\mT_{3}^{t}$ using the CLIP tokenizer. The output from the CLIP text encoder, enriched with information from the original point cloud, is instrumental in enabling effective text-to-3D generation. The final output is achieved using Point-E. This framework is inspired by BLIP-Diffusion~\cite{blip-diffusion} techniques used in subject-driven 2D generation. However, the key distinction here from BLIP-Diffusion lies in the way we concatenate the Clip text token and Q-Former feature. This difference may also stem from variations in the data volumes between 2D and 3D, which will be thoroughly examined in the appendix.

\newpage
\begin{figure*}[!t]
    \centering
    \includegraphics[width=0.93\linewidth]{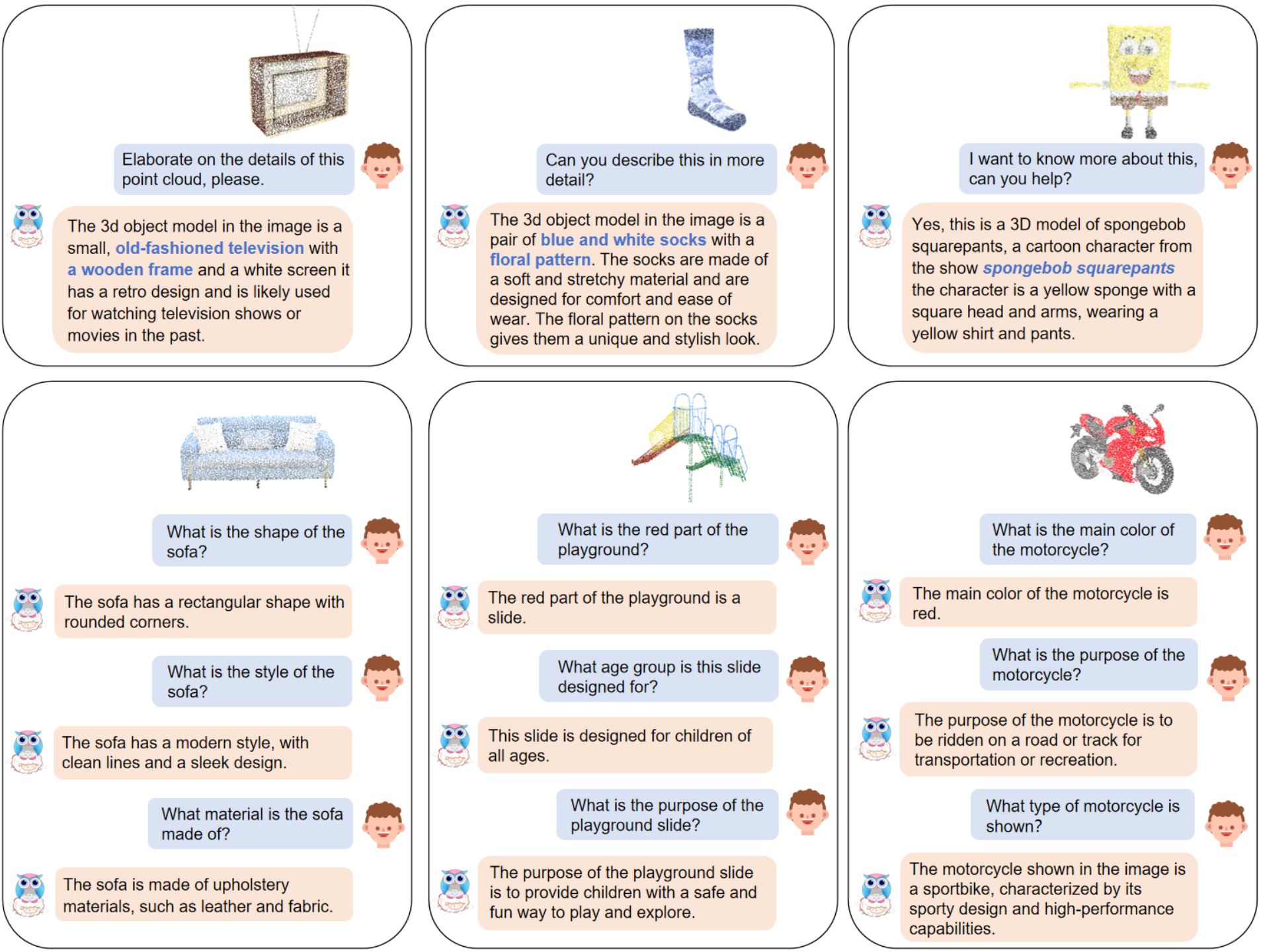}
    \vspace{-2mm}
    \caption{\textbf{Examples of text inference using the GPT4Point with ViT-g and OPT6.7B after Instruct Finetuning.} The table showcases its proficiency with point cloud input, excelling in tasks like detailed caption generation and point cloud-based question answering. This underscores our model's profound grasp of point cloud geometry and color, translating them into meaningful semantics.}
    \label{fig4_qa_demo}
    \vspace{-5mm}
\end{figure*}

\section{Benchmarks and Evaluation}
\label{sec_4 Benchmarks and Evaluation}
~~Evaluating the performance of multimodal models presents significant challenges due to the lack of mature metrics for assessing the quality of generated texts. For 3D objects, benchmarks primarily rely on human judgment or GPT-based assessments~\cite{Pointllm}. There are two key issues to consider in this context. Firstly, the evaluation process involves a certain degree of subjectivity. Identical results might receive varying scores, leading to an element of randomness. Secondly, each evaluation incurs time and monetary costs. In this section, we present the evaluation benchmark we have proposed, which is primarily designed to be objective, ensuring repeatability and verifiability. \cref{sec_4.1. Composition of Test Set} outlines the composition of our test set. \cref{sec_4.2. 3D Object Recognition} addresses the evaluation of recognition capabilities, while \cref{sec_4.2. 3D Object Recognition} provides a detailed assessment of text inference abilities.

\subsection{Composition of Test Set}
\label{sec_4.1. Composition of Test Set}
~~We leverage the Objaverse dataset~\cite{Objaverse}, aligning it with LVIS categories~\cite{LVIS}, to create Objaverse-LVIS validation and test sets. In Objaverse-LVIS, we exclude scenes with complex settings, such as indoor houses or outdoor parks, focusing more on scenarios with single objects or combinations of multiple objects. We construct validation and test sets, each containing 1K objects. Compared to the PointLLM~\cite{Pointllm}, which uses only 200 unfiltered objects as a test set, our larger set of 1K objects better measures the model's generalization capabilities. For textual descriptions, we initially use Pyramid-XL to get initial annotations, followed by multiple rounds of expert manual revisions, ensuring comprehensive and accurate descriptions.

\subsection{3D Object Recognition}
\label{sec_4.2. 3D Object Recognition}
~~~~3D object recognition represents the classification capabilities of 3D multimodal models and the ability to match point cloud features with textual features. Objective measures, like accuracy, are typically used for evaluation.

\vspace{2pt}
\noindent \textbf{Zero-shot point classification.} 
Zero-shot point classification is considered a classic task in this domain. The widely used ModelNet40 dataset~\cite{ModelNet40}, which includes 2,468 objects across 40 categories, serves as a benchmark to evaluate a model's classification capabilities. In the multimodal context, the typical approach involves using the text \textit{'a 3D model of [name]'} as input to match with the point cloud modal features. The accuracy metric ACC@1, indicating the precision of top-1 rankings, best reflects the model's ability to accurately match object categories.

\vspace{2pt}
\noindent \textbf{3D point-text retrieval.}
In 3D Point-Text Retrieval, we initially select 128 candidates based on point-text feature similarity and then re-rank these candidates using matching scores. Unlike classification tasks where the text usually involves simple category names, here the text can be more complex descriptions. The evaluation metrics used are similar to those in image-text retrieval. We employ R1, R5, and R10 metrics to measure the accuracy of the top 1, 5, and 10 results in correctly matching points to text and vice versa.

\subsection{3D Object Text Inference}
\label{sec_4.3 3D Object Text Inference}
~~~~3D object text inference deeply represents the understanding capabilities regarding objects, including 3D object point cloud captioning and 3D point cloud question answering.

\vspace{2pt}
\noindent \textbf{3D point cloud captioning.}
This task primarily evaluates the model's ability to provide an overall summary of a 3D object. The captions in the Objaverse-XL-LVIS caption test set are mostly within 30 words and accurately describe the object's geometry, color, and state. And we predominantly employ common image description metrics, such as BLEU1, BLEU4, METEOR, ROGUE-L, and CIDEr~\cite{BLEU, ROUGE, METEOR} for evaluation.

\vspace{2pt}
\noindent \textbf{3D point cloud question answering.}
In addition to point cloud captioning, 3D point cloud question answering explores object details through multiple rounds of dialogue. For instance, we can further explore the color or shape of specific parts of an object or even infer its simple usage. The curated Objaverse-XL-LVIS short QA 1K test set features concise, straightforward questions and answers, allowing us to conveniently calculate answer accuracy. Besides accuracy, we also use metrics from captioning to evaluate model performance. It is important to note that, for a fair comparison, we solely utilize zero-shot learning, meaning no fine-tuning is conducted on this kind of short QA dataset.

\begin{table*}[!t]
    \small
	\centering	
 \setlength\tabcolsep{4pt}
	\resizebox{0.90\textwidth}{!}{%
	\begin{tabular}	{l >{\raggedright\arraybackslash}p{2.5cm} |  c  c  c  c  c  c |  >{\centering\arraybackslash}p{2.5cm}  }
		\toprule	 	
	 \multirow{3}{*}{Model} & \multirow{3}{*}{Input Data Type} & \multicolumn{6}{c|}{ObjaverseXL-LVIS \textbf{Retrieval} (1K test set)}   & ModelNet40\cite{ModelNet40}\\
	 & &    \multicolumn{3}{c}{Point $\rightarrow$ Text}& \multicolumn{3}{c|}{Text $\rightarrow$ Point} & Accuracy\\
	     
	& &  R@1&R@5&R@10& R@1&R@5&R@10& Acc@1 \\
	\midrule  
    \footnotesize{\textit{~~Image-Text Modal}} &  \\	   
	\footnotesize{BLIP-2} & \multirow{2}{2.5cm}{{\fontsize{8}{6}\selectfont Single-View Image \\ (Mesh with Color)}}
	& 17.56 & 41.16 & 52.82 & 16.72 & 40.2
	& 52.56  & 35.62
	\\
	  \footnotesize{InstructBLIP\textsuperscript{†}} & 
	& 20.4 & 43.1 & 55.3 & 13.7 & 32.5 & 42.7 
	& 31.48 \\
	\midrule
	\multicolumn{2}{l}{\footnotesize{\textit{~~Point-Text Modal}}}
	\\
    \footnotesize{PointLLM}\scriptsize{(Vicuna-7B)\textsuperscript{†}} &  \multirow{2}{2.5cm}{{\fontsize{8}{6}\selectfont Point Cloud (+Color)}}
    & - & - & - & - & - & -
    & 41.33 \\    
    \footnotesize{\textbf{GPT4Point}} & 
     & \textbf{32.2} & 64 &\textbf{81.3}  &  \textbf{89.7}& \textbf{98.1} &\textbf{98.9}
    & \textbf{43.90}  \\  
    \bottomrule
	\end{tabular}
	}
  \vspace{-4pt}
	\caption
	{	
        \textbf{Point-Text Retrieval on the Objaversexl-LVIS test dataset and zero-shot 3D classification on ModelNet40}. Please note that \textsuperscript{†} denotes Generative 3D object classification, which refers to the process of classifying 3D objects based on the generation of captions.
	}
	\label{t1_retrieval}
 \vspace{-1ex}
\end{table*}

\begin{table*}[!t]
    \small
	\centering	
	\setlength\tabcolsep{4pt}
	\resizebox{0.90\textwidth}{!}{
	\begin{tabular}{l l | c c c c c | >{\centering\arraybackslash}p{1.1cm} >{\centering\arraybackslash}p{1.1cm} >{\centering\arraybackslash}p{1.3cm}} 
		\toprule	 	
	 \multirow{2}{*}{Model} & \multirow{2}{*}{\makecell[l]{\#Trainable \\ Params}} & \multicolumn{5}{c|}{ObjaverseXL-LVIS \textbf{Caption} (1K test set)} & \multicolumn{3}{c}{ObjaXL-LVIS \textbf{QA} (1K)}\\
	     
	& & BLEU1 & BLEU4 & METEOR & ROUGH-L & CIDEr & \multicolumn{1}{c|}{\textbf{Acc}} & BLEU1 & ROUGH \\
	\midrule  
    \textit{~~Image-Text Modal} \\	   
	BLIP-2 (OPT\textsubscript{2.7B}) & 188M
	& 22.2 & 3.0 & 10.3 & 28.2 & 32.3 & \multicolumn{1}{c|}{13.4} & 14.2 & 16.8
	\\
 	BLIP-2 (OPT\textsubscript{6.7B}) & 188M
	& 24.9 & 4.1 & 11.5 & 30.0 & 44.2 & \multicolumn{1}{c|}{15.4} & 15.1 & 18.3
	\\
	  InstructBLIP(Vicuna-1B) & 202M
	& 25.5 & 4.3 & 11.6 & 30.7 & 47.2 & \multicolumn{1}{c|}{15.9}
	& 16.2 & 20.1 \\
	  Qwen-VL(Qwen-7B) & 7.2B
	& 27.1 & 4.9 & 13.1 & 31.3 & 63.8 & \multicolumn{1}{c|}{18.2}
	& 19.5 & 24.4 \\
	\midrule
    \textit{~~Point-Text Modal} \\
    PointLLM (Vicuna-13B)\textsuperscript{†} & 13.3B 
    & 26.2 & 4.9 & 11.9 & 31.3 & 50.9 & \multicolumn{1}{c|}{23.4}
    & 22.3 & 26.2
    \\    
    \textbf{GPT4Point} (OPT\textsubscript{2.7B}) & 110M
    & 28.9 & 6.0 & 13.2 & 33.9 & 68.4
    & \multicolumn{1}{c|}{22.1} & 23.4  & 25.3
    \\
    \textbf{GPT4Point} (OPT\textsubscript{6.7B}) & 110M
     & 31.5 & \textbf{7.2} & 13.8 & 35.4 & \textbf{78.7}
    & \multicolumn{1}{c|}{27.1} & 26.2 & 30.4
    \\  
    \textbf{GPT4Point} (FLANT5\textsubscript{XL}) & 110M
     & \textbf{32.2} & \textbf{7.2}& \textbf{14.2}  & \textbf{35.5}& 78.0
    & \multicolumn{1}{c|}{\textbf{27.6}} & \textbf{26.3} & \textbf{31.3}
    \\      
    \bottomrule
	\end{tabular}}
  \vspace{-2pt}
	\caption
	{	
        \textbf{3D Object Point Caption and Question Answer (QA) on the Objaversexl-LVIS 1K test dataset.} For the BLIP series, only fine-tuning of the Q-Former structure is required, whereas models like PointLLM need fine-tuning of the large language model.
	}
	\label{t2_caption_and_qa}
 \vspace{-8pt}
\end{table*}

\section{Experiments}
\subsection{Training Details}
~~We configure our setup to process 8,192 input point clouds, utilizing Point-BERT~\cite{Pointbert} as the backbone. This transformer-based network excels in capturing geometric and semantic features of object point clouds. And the backbone is pretrained through retrieval tasks like ULIP-2~\cite{ulip2}. We employ OPT~\cite{OPT} and FlanT5~\cite{flan, t5} as Large Language Models (LLMs). For the training process, we adopt an initial learning rate of 1e-4, weight decay of 0.05, batch size of 32, and the AdamW optimizer~\cite{adamw}. All hyperparameters remain unchanged in both stages. The training process takes 10 epochs for each stage on 8 A100 GPUs.

\subsection{Evaluation and Diverse Tasks}
\label{sec_4.2. Evaluation and Diverse Tasks}
~~~~We evaluate our model on the benchmark we proposed in \cref{sec_4 Benchmarks and Evaluation}, which includes 3D object recognition and 3D object text inference. Additionally, we demonstrate the model's capability for controllable text-to-3D generation.

\vspace{2pt}
\noindent \textbf{3D object recognition.}
Recognition capabilities are shown in \cref{t1_retrieval}, with zero-shot classification results on the right side. Our approach demonstrates superior performance, outperforming the Vision Language Model(VLM) InstructBLIP~\cite{InstructBLIP} by 12.42 points and surpassing PointLLM~\cite{Pointllm} by 2.57 points. Notably, PointLLM employs a generative approach to generate the text results by a prompt, limiting its direct recognition capabilities. The results for 3D point-text retrieval are shown on the left side. Our GPT4Point model outperformed other VLMs~\cite{BLIP-2, InstructBLIP, Qwen-VL}. The results quantitatively highlight the challenges of single-viewpoint 3D object occlusions and biases, emphasizing our approach's advantages over other image-text models.

\vspace{2pt}
\noindent \textbf{3D object text inference.}
Model's text inference capabilities are displayed in \cref{t2_caption_and_qa}. On the left, the results of 3D object point cloud captioning confirm GPT4Point's superiority over pretrained VLMs and PointLLM. Notably, the Point Q-Former structure allows freezing the LLM, significantly reducing training parameters. The results for 3D point cloud question answering on the right side show that GPT4Point achieved the best zero-shot accuracy, surpassing InstructBLIP~\cite{InstructBLIP} by 11.7 points and outperforming PointLLM~\cite{Pointllm} by 4.2 points. Alongside quantitative results, \cref{fig4_qa_demo} qualitatively demonstrates its detailed answers and multi-turn dialogue capabilities, with more examples in the appendix.

\begin{figure*}[!t]
    \centering
    \includegraphics[width=0.9\linewidth]{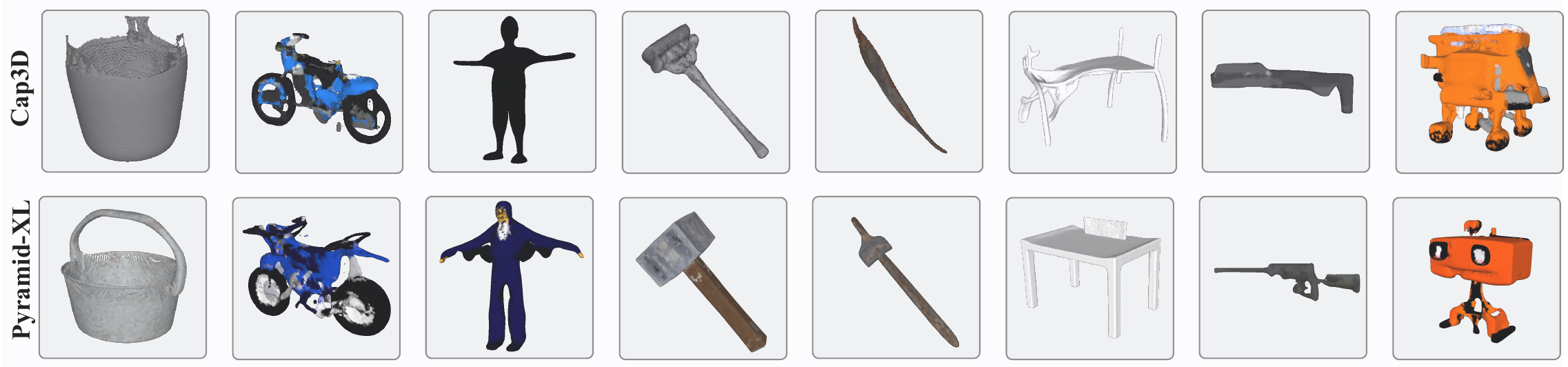}
    \caption{\textbf{Object generated from Point-E fine-tuned on Cap3D~\cite{Cap3D} and our Pyramid-XL} The first line shows Cap3D~\cite{Cap3D} fine-tuning results, while the second, using our Pyramid-XL Level 3 Dense Caption, outperforms Cap3D in geometry and color. This underscores the high quality of our text annotations. }
    \label{fig5_dataset_pointe_ft_results}
    \vspace{-12pt}
\end{figure*}

\begin{figure}[!t]
    \centering
    \includegraphics[width=0.9\linewidth]{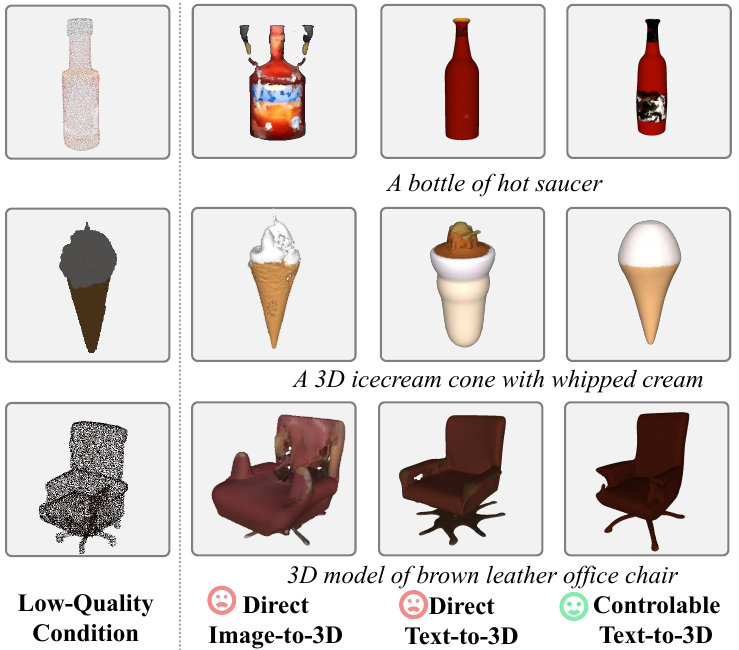}
    \caption{\textbf{Point-E generation result when conditioned on text, single image and our Point-Q-former features}}
    \label{fig6_point_diffusion}
    \vspace{-4ex}
\end{figure}

\vspace{2pt}
\noindent \textbf{Controllable text-to-3D object generation.}
Here, we showcase the generative capabilities of our model. Given features of low-quality point clouds along with textual descriptions, we can generate corresponding higher-quality point clouds, making text-to-3D more controllable. \cref{fig6_point_diffusion} displays experimental results,  We compare our point feature condition with text or single image condition in Point-E, demonstrating that aligning features using both point cloud and textual information significantly improves guidance for point cloud generation. It is worth noticing that when compared to a single view image rendered from the original 3D model, our Point Q-former feature serves as a better condition that contains richer information about the geometric shape and detailed color information of 3D objects. We believe this is the first step towards the point cloud editing.

\subsection{Assessing the Effectiveness of Pyramid-XL}
\label{sec_5.3_Assessing the Effectiveness of Pyramid-XL}
~~~In this section, we demonstrate the effectiveness of Pyramid-XL in obtaining high-quality point-text annotations. We focus on two tasks: fine-tuning Point-E~\cite{Point-E} for 3D object generation using dense captions and utilizing annotations of varying granularities on the QA benchmark.

\vspace{2pt}
\noindent \textbf{Finetune the Point-E with Level 3 Caption.}
We fine-tuned Point-E~\cite{Point-E} base-40M text-vec model using 70K Level 3 VLM instruction captions from Pyramid-XL for 3D object generation. The results in~\cref{fig5_dataset_pointe_ft_results} show significant improvements in geometric details and and color fidelity in point clouds, especially in objects like baskets and Halloween costumes, compared to Cap3D~\cite{Cap3D}.

\begin{table}[!t]
	\centering	
    \setlength\tabcolsep{4pt}
	\resizebox{0.89\columnwidth}{!}{%
	\begin{tabular}	{l  |  c  c | c  }
		\toprule	 	
	 \multirow{2}{*}{\small{Text-to-3D methods}}&  \multicolumn{2}{c}{\small{Rendering Eval}} & \small{User Study} \\
	 & \footnotesize{FID} \(\downarrow\) & \footnotesize{CLIP Score} & \footnotesize{Score(1-5)} \\
		\midrule
        \footnotesize{Direct text-to-3D} &   \footnotesize{34.7} &  \footnotesize{74.9} &   \footnotesize{3.98}  \\
        \footnotesize{Direct image-to-3D} &   \footnotesize{32.6} &  \footnotesize{75.3} &   \footnotesize{3.67}  \\
         \footnotesize{Controllable text-to-3D} &   \footnotesize{\textbf{31.6}} &  \footnotesize{\textbf{76.2}} &   \footnotesize{\textbf{4.03}} \\
	\bottomrule	
	\end{tabular}
 	}
 \vspace{-1ex}
	\caption
	{
	\small	
        Different 3D generation methods on the Cap3D, 2K test set. Our controllable text-to-3D achieved the best results.
	}
	\vspace{-2ex}
	\label{t3_generation_results}
\end{table}		 
\begin{table}[!t]
	\centering	
    \setlength\tabcolsep{4pt}
	\resizebox{0.85\columnwidth}{!}{%
	\begin{tabular}	{l  |  c  c  }
		\toprule	 	
	 \multirow{2}{*}{\small{Level of \textit{Pyramid-XL}}}&  \multicolumn{2}{c}{\small{3D Object QA}} \\
	 & val & test \\
		\midrule
         \footnotesize{Level 2} &   \footnotesize{22.3} &  \footnotesize{22.1}\\
         \footnotesize{Level 1 + Level 2} &   \footnotesize{25.6} &  \footnotesize{25.4} \\
         \footnotesize{Level 1 + Level 2 + Level 3 (30\%)}  &  \footnotesize{27.3} &  \footnotesize{27.1} \\
		 \footnotesize{Level 1 + Level 2 + Level 3 (70\%)} &  \footnotesize{28.3} &  \footnotesize{28.2} \\
        \footnotesize{Level 1 + Level 2 + Level 3 (100\%)} &  \footnotesize{\textbf{28.5}} &  \footnotesize{\textbf{28.4}} 
       \\
	\bottomrule	
	\end{tabular}
 	}
 \vspace{-1ex}
	\caption
	{
	\small	
        \textbf{the impact of text granularity.} Levels 1, 2, and 3 represent coarse single-view annotation, multi-view annotation, and fine-grained annotation, respectively. We utilize the large language model OPT2.7B for pretraining and evaluation using ObjaverseXL-LVIS validation and test sets after finetuning.
	}
	\vspace{-2ex}
	\label{t4_pretrain_finetune}
\end{table}		 

\vspace{2pt}
\noindent \textbf{Ablation study in model pretraining.}
Our ablation studies on Pyramid-XL, detailed in \cref{t4_pretrain_finetune}, investigated the impact of pretraining data scale and quality on model performance. The comparison between the first two rows indicates that using a large volume of coarse annotations boosts baseline performance. Additionally, incorporating a higher proportion of detailed Level 3 annotations leads to improved QA scores, with 80\% yielding near-optimal results.

\section{Conclusion}
We introduce the innovative GPT4Point, a Unified Framework for point-language understanding and generation including the 3D MLLM for point-text tasks and controlled text-to-3D generation based on low-quality point feature.
We develop Pyramid-XL, a point-language dataset annotation engine. This setup constructs a large-scale database over 1M objects of varied coarseness levels from the Objaverse-XL dataset.
Furthermore, we establish an object-level point cloud benchmark with specific metrics for evaluating 3D point cloud-language tasks. This benchmark provides a comprehensive approach to assess both the understanding abilities of 3D multimodal language model and the quality of generated objects.

\clearpage
\newpage
\renewcommand{\thetable}{S\arabic{table}}
\renewcommand\thefigure{S\arabic{figure}}
\setcounter{figure}{0}
\setcounter{table}{0}

\vspace{20pt}

\appendix
\section{Supplementary Material Introduction}
~~~In this supplementary material, we extend the discussions presented in the main conference paper. Sec.~\ref{sup_sec_B Additional Related Work} provides a more in-depth exploration of related work, focusing on defining the scope of large language models family and examining the developments in point-text multimodal approaches. Sec.~\ref{sup_sec_C Additional Method} supplements more details about the data annotation engine: Pyramid-XL and the diffusion architecture. Moving to Sec.~\ref{sup_sec_D Additional Benchmark}, we expand on the superiority of our benchmark. Initially, we introduce examples from our ObjaverseXL-LVIS QA 1K dataset, which includes concise QAs for evaluation and long QAs for instructive tuning. Then we show more 3D generation failure cases where GPT4Point can figure it out while 2D VLM can not to underscore the necessity and relevance of our 3D point-text benchmark. Finally in Sec.~\ref{sup_sec_E Additional Experiments}, we give more qualitative results of Point-text inference tasks including caption and QA tasks and Controllable point diffusion. 

\section{Additional Related Work}
\label{sup_sec_B Additional Related Work}
~~~In this section, we provide detailed insights into related work. Sec.~\ref{sup_sec_B_1 The Family of LLMs and MLLMs} classifies key concepts of large language models, including LLMs, MLLMs, and VLMs. Sec.~\ref{sup_sec_B_2 Additional The development of Point-text Multimodal} present the evolution of point-text multimodal models through an illustrative flowchart.

\subsection{The Family of LLMs and MLLMs}
\label{sup_sec_B_1 The Family of LLMs and MLLMs}
\begin{figure}[!h]
    \centering
    \includegraphics[width=0.92\linewidth]{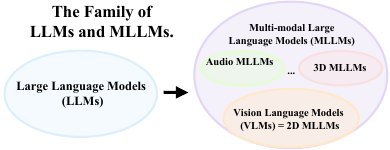}
    \vspace{0.2mm}
    \caption{\textbf{The Family of LLMs and MLLMs.} }
    \label{sup_fig1_LLMs_VLMs_MMLMs}
    \vspace{-2mm}
\end{figure}

Although the concepts related to large language models are already familiar, we still wish to detail these concepts here. We briefly introduce some families of LLMs and MLLMs. First are the LLMs based on the Transformer architecture, such as ChatGPT~\cite{ChatGPT} and GPT-4~\cite{GPT4}. Currently, there are several open-source, deployable models~\cite{llama, vicuna, OPT, flan}. After extensive pre-training on a vast corpus, they exhibit strong comprehension and reasoning abilities. Multimodal Large Models (MLLMs) aim to enable LLMs to understand information in other modalities. The fundamental approach involves retrieving text features with other modality features. Among them, image-text multimodal large models, also known as 2D MLLMs or Visual Language Models (VLMs), stand out due to the abundant image-text pairs and strong image backbones provided by computer vision~\cite{BLIP-2, LLaVA}. Beyond images, there are other modalities, such as Audio MLLMs~\cite{Audiogpt} that combine with the audio modality and Video MLLMs with the video modality~\cite{VideoLLM}. In the 3D domain some existing work, like 3D-LLM~\cite{3d-llm}, utilizes 2D image features combined with depth projections to generate 3D features. We propose a unified text understanding and generation model based on point clouds and develop a real 3D MLLM.

\subsection{The development of Point-text Multimodal}
\label{sup_sec_B_2 Additional The development of Point-text Multimodal}
\begin{figure}[!h]
    \centering
    \includegraphics[width=\linewidth]{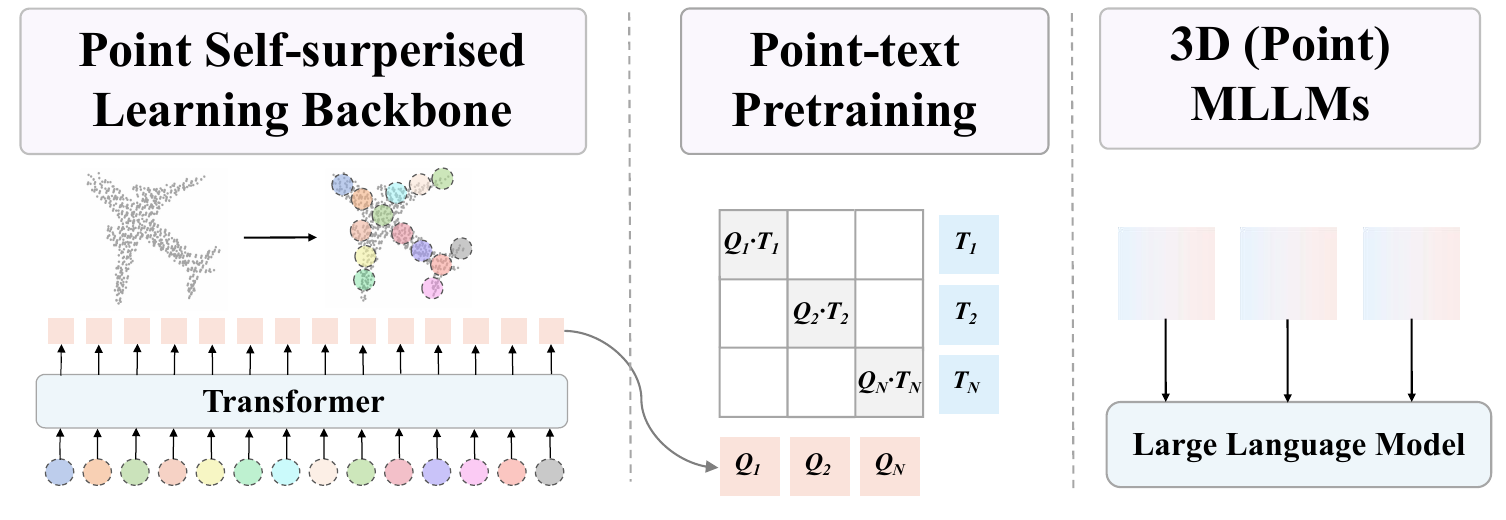}
    \vspace{-2mm}
    \caption{\textbf{The development of Point-text Multimodal.} }
    \label{sup_fig2_point_text_development}
    \vspace{-2mm}
\end{figure}

In this section, we delve into the evolution of point-text multimodal models for single objects.
\begin{itemize}
\item \textbf{Backbone Development:} The foundational aspect of our methodology lies in the robust development of the backbone for handling point clouds. Similar to the methodologies applied to texts and images, point clouds undergo a self-supervised training strategy to establish a strong foundation. Notably, we leverage the innovative PointBert~\cite{Pointbert} framework, which divides point clouds into patches and executes a reconstruction process on masked patches. This is achieved through the utilization of a Transformer-based backbone, imparting a powerful and adaptive feature extraction capability to our model.

\item \textbf{Text Modality Alignment:} Drawing inspiration from the successful model CLIP~\cite{CLIP}, our approach incorporates a phase dedicated to aligning point patches with textual features. This strategic alignment augments the backbone's inherent ability to process textual information seamlessly. By fusing the spatial understanding of point clouds with the semantic richness of textual data, our GPT4Point achieves a more comprehensive and nuanced representation, enhancing its overall performance.

\item \textbf{3D MLLMs Integration:} Building upon the successful alignment of point patches with textual features, the next crucial step involves the integration of point features into Large Language Models (LLMs). This integration mirrors approaches seen in Vision Language Models (VLMs) and extends their capabilities to comprehend and interpret point cloud data. The seamless fusion of 3D spatial information with the linguistic context empowers Large Language Models (LLMs) with a more holistic understanding of the data, enabling them to discern intricate patterns and relationships within the point clouds.

\end{itemize}


\newpage
\section{Additional Method}
\label{sup_sec_C Additional Method}
~~~Here, we provide additional information to our method. We first give more details about the data text annotation engine Pyramid-XL in Sec.~\ref{sup_sec_3_1 Pyramid-XL: Data Annotation Engine}. And then, in Sec.~\ref{sup_sec_C_2 Point Diffusion Architecture} about the model architecture, we give the details about the point diffusion branch.

\subsection{Pyramid-XL: Data Annotation Engine}
\label{sup_sec_3_1 Pyramid-XL: Data Annotation Engine}
~~~First, we introduce the approach to acquire point cloud from Objaverse-XL~\cite{Objaverse-XL}. Then we introduce the cost and prompts of the our data annotation engine Pyramid-XL. Finally, we give more qualitative results that finetune the Point-E~\cite{Point-E} by our Pyramid-XL level 3 dense captions.

\begin{figure}[h]
    \centering
    \includegraphics[width=\linewidth]{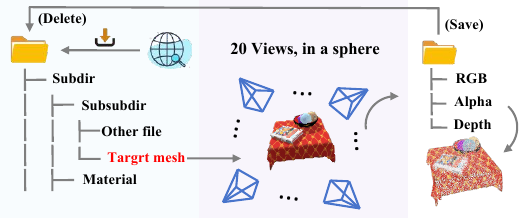}
    \vspace{0.2mm}
    \caption{\textbf{Acquire Data Pipeline from Objaverse-XL~\cite{Objaverse-XL}.}}
    \label{sup_fig3_objaversexl_pipeline}
    \vspace{-2mm}
\end{figure} 

\mypara{Acquire data from Objaverse-XL.}
Here we detail our processing approach for the Objaverse-XL dataset~\cite{Objaverse-XL}. It has 10M objects and is the extension of Objaverse-1.0~\cite{Objaverse} which only has 800K 3D objects. Objaverse-XL offers only unprocessed downloads for its 3D objects, most of which originate from sources like GitHub. Downloading these mesh files necessitates obtaining the complete project, as materials and related components are often stored in other separate directories. Downloading the raw dataset in this format is impractical due to excessive memory requirements, with an average project consuming about 1GB of space. Therefore, we render object images and clear the cache upon completion to manage space. We render 20 random views of each object, capturing the RGB, alpha values, and depth, which are then used to generate point clouds. In addition to the 780K objects from Objaverse-1.0, we rendered an additional 220K from Objaverse-XL, totaling 1M objects.

\begin{table}[!ht]
\centering
\begin{adjustbox}{scale=0.75}
\begin{tabular}{@{}lccccc@{}}
\toprule
Dataset & Num Obj & Data Type & Cost/K (GPU + GPT) \\
\midrule
Level 1 & 1M &  Single-View Caption  & \$0.47 + \$0 \\
Level 2 (GPT-4) & 660K &  Multi-View Caption & \$4.17 + \$4.18* \\
Level 2 (ChatGPT) & 660K &  Multi-View Caption & \$4.17 + \$0.14* \\
Level 3 & 70K & QA, Detailed Caption & \$1.64 + \$0  \\
\bottomrule
\end{tabular}
\end{adjustbox}
\caption{
\textbf{Comparing Costs across Different Dataset Levels.} Costs are calculated based on generating annotation for 1K objects. * is directly from Cap3D~\cite{Cap3D}. As levels increase, the cost rises, indicating larger scales for lower-level datasets.
}
\label{sup_tab1_Pyramid-XL_cost}
\vspace{-3.5mm}
\end{table}
\mypara{The cost of the Pyramid-XL.}
We now turn our attention to the cost analysis of our data annotation engine, detailed in~\cref{sup_tab1_Pyramid-XL_cost}. The primary costs, detailed under the '1K Cost' column, include GPU resources on the left and GPT API usage on the right. We use the same GPU settings as Cap3D~\cite{Cap3D}, employing A40s on a identical cloud platform. Given GPUs' parallel processing, costs are equal for single or multiple units. We calculate usage time assuming a single GPU for simplicity. For Level 1, we use BLIP-2~\cite{BLIP-2} to generate one short caption for one object. It needs 0.074 hours and costs $0.074h \times \$1.28/h = \$0.095$. For Level 2 the cost is the same as the Cap3D~\cite{Cap3D}. The GPU resource fees include BLIP-2~\cite{BLIP-2} and CLIP~\cite{CLIP}. BLIP-2 generates 8 views for each object and each view has 5 captions, so the fee is $\$0.095 \times 8 \times 5 = \$3.76$. And the CLIP uses 0.3h and costs $0.3h \times \$1.28/h = \$0.38$. All GPU resource fee is $\$3.76 + \$0.38 = \$4.17$. For the GPT API fee, it costs \$0.03/1k tokens and needs 139.3 tokens for each object and the total cost is $\$139.3/1000k \times \$0.03/1k \times 1000 = \$4.18$. For Level 3, We use the open-source Visual Language Model (VLM) Qwen-VL~\cite{Qwen-VL} for processing the final captions. It needs 1.28h for the CLIP filter and Qwen-VL generation captions, so the cost is $1.28h \times \$1.28/h = \$1.64$. 
 
We can observe that Level 2 captions account for most of the costs, primarily due to GPT usage fees. Our findings show that using GPT-4 for text-based multi-view caption synthesis doesn't substantially outperform ChatGPT. Furthermore, by utilizing open-source Large Language Models (LLMs), we can entirely eliminate API call expenses. The other major cost is the GPU resources, as it uses BLIP-2 to generate five captions for each view, which can lead to redundancy in information. We can reduce the number of captions for each view, and even the number of views.

\mypara{The prompts of the Pyramid-XL.}
We present the prompt part of the Pyramid-XL data text annotation engine, as illustrated in~\cref{sup_fig6_prompt1} and~\cref{sup_fig7_prompt2}. We primarily focus on illustrating how to construct GPT-based Level 2 captions, ChatCaptioner-based Level 3 short QA pairs, and MLLM-based Level 3 instruction captions and long QA pairs. 

For Level 2 captions, we use Level 1 captions of rendered images from 6 views. Through carefully designed prompts, we integrate captions from the 6 captions to obtain a comprehensive and relatively accurate caption with fewer than 30 words. In our paper, we use GPT-4 to get the comprehensive caption but we find that ChatGPT can be replaced by GPT-4 to generate Level 2 captions to reduce the cost.

For Level 3 short QA, we follow the approach outlined in ChatCaptioner~\cite{ChatCap}. We use ChatGPT or other LLMs (we choose Vicuna-7B~\cite{vicuna}) as the questioner and BLIP-2~\cite{BLIP-2} as the answerer. By providing appropriate instructions and context (Level 2 caption) to both the LLM and BLIP-2, we observe that, LLM generate diverse questions that that include aspects such as color, type, material, purpose, and more. Also, without restricting the number of words, BLIP-2 tends to output concise answers. These form the basis for our Objaverse-XL short QA dataset.

For Level 3 dense captions, we use the Level 2 caption as context, feed the rendering image that best matches the context into MLLM, and input suitable instructions. Due to a combination of high-quality conversational performance and cost-effectiveness, we choose the Qwen-VL~\cite{Qwen-VL} model to generate. The construction method for Level 3 instruction (long) QA pairs is similar to the above steps, with the key difference lying in the variation of instructions.

\mypara{The effectiveness of Pyramid-XL Level 3 caption.}
We use dense captions from Level 3 of Pyramid-XL to fine-tune Point-E and compare the results with those of Cap3D, as shown in~\cref{sup_fig11_dataset_pointe_ft_results}. Ours significantly outperform Cap3D's captions, demonstrating the precision of our captions.

\subsection{Point Diffusion Architecture}
\label{sup_sec_C_2 Point Diffusion Architecture}

\begin{figure}[h]
    \centering
    \includegraphics[width=\linewidth]{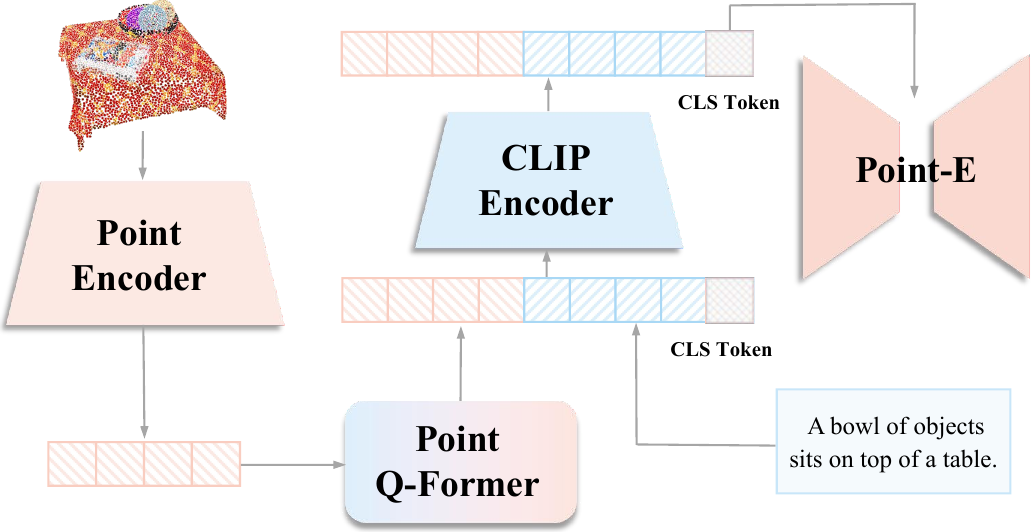}
    \caption{\textbf{Point Diffusion with GPT4Point.} }
    \label{sup_fig4_point_diffusion}
    \vspace{-4mm}
\end{figure}

Currently, there are indeed some explorations into controllable text-to-3D work~\cite{TextMesh, DreamTime, JM3D}. However, we are attempting to combine understanding and controllable 3D generation together.
Here, we offer an in-depth look at the Diffusion branch's structure in Stage 2, illustrated in~\cref{sup_fig4_point_diffusion}. Initially, the point cloud undergoes processing via the Point Encoder (Backbone) and Point Q-Former, yielding Q-Former Tokens. For text, instead of the Point Q-Former's text tokenizer, we utilize Point-E's CLIP tokenizer. The resulting text tokens are then concatenated with the Q-Former Tokens. Subsequently, the CLS token from the Text Token is fed into Point-E. The concatenation method in GPT4Point differs notably from BLIP-Diffusion~\cite{blip-diffusion}. In BLIP-Diffusion, Q-Former Tokens are inserted between the CLS token and input tokens. In contrast, GPT4Point appends Q-Former Tokens directly to the text token sequence, allowing the CLS token to integrate both geometric and color information, crucial for guiding the 3D generation.

\section{Additional Benchmark}
\label{sup_sec_D Additional Benchmark}
~~~In this section, we mainly introduce some additional contents about the benchmark. In Sec.~\ref{sup_sec_D_1 ObjaverseXL-LVIS QA 1K dataset}, we give more examples of the ObjaverseXL QA dataset. Note that the short QA dataset is used for evaluation based on the accuracy metric. Then in Sec.~\ref{sup_sec_D_2 Anomalous Objects: Generation Failure Cases}, we show more qualitative results about Generation Failure Cases which can not be recognized by 2D VLMs through a single view but are judged by our GPT4Point.

\subsection{Objaverse-XL QA  Dataset}
\label{sup_sec_D_1 ObjaverseXL-LVIS QA 1K dataset}

\mypara{Short QA Dataset} 
We use the short QA dataset for the evaluation of the 3D point cloud question answering task. We selecte categories that overlap with both Objaverse-XL and LVIS~\cite{LVIS}, constructing 1K Point-QA data as the test set. The specific samples are presented in~\cref{sup_fig8_qa_dataset}, which includes questions covering various aspects such as color, material, composition, category, etc. The answers are concise, with an average word length of 2.32, making them convenient for testing. We use accuracy top-1 as metric and evaluate the model's zero-shot short QA capability on this dataset.

\mypara{Long (Instruction) QA Dataset}
The long (Instruction) QA dataset is for finetuning the model to significantly enhance the model's conversational capabilities. We impose length constraints on prompts, requiring approximately 50 words for answers to dense caption questions and not less than 10 words for other questions. As illustrated in~\cref{sup_fig8_qa_dataset}, we constructed a Long (Instruction) QA dataset for 70K objects, comprising 344,996 QA pairs. Among these, 69K data are used for fine-tuning, while the remaining 1K are reserved for testing. This aims to encourage LLMs to generate long and more comprehensive results.

\subsection{Anomalous Objects: Generation Failure Cases}
\label{sup_sec_D_2 Anomalous Objects: Generation Failure Cases}
~~~In this section, we will demonstrate more qualitative results to show the failure case which can not be recognized by 2D VLMs through a single view but can be judged by our GPT4Point. In this section, we mainly show the failure cases produced by the state of the arts text-to-3D generation methods like Dream-Gaussian~\cite{dreamgaussian} and Fantasia3d~\cite{Fantasia3d}. Due to technical constraints, these models are likely to generate 3D objects with multi-heads or multi-bodies. If provided with render images from only a single perspective, 2D VLMs~\cite{Qwen-VL, vicuna}, and even humans in most cases, may make incorrect judgments, as illustrated in the upper part of~\cref{sup_fig9_qa_special}. This hinders the assessment of 3D object generation. However, our GPT4Point provides a better solution to this issue. More examples are showcased in \cref{sup_fig9_qa_special}.

\begin{figure*}[!t]
    \centering
    \includegraphics[width=0.91\linewidth]{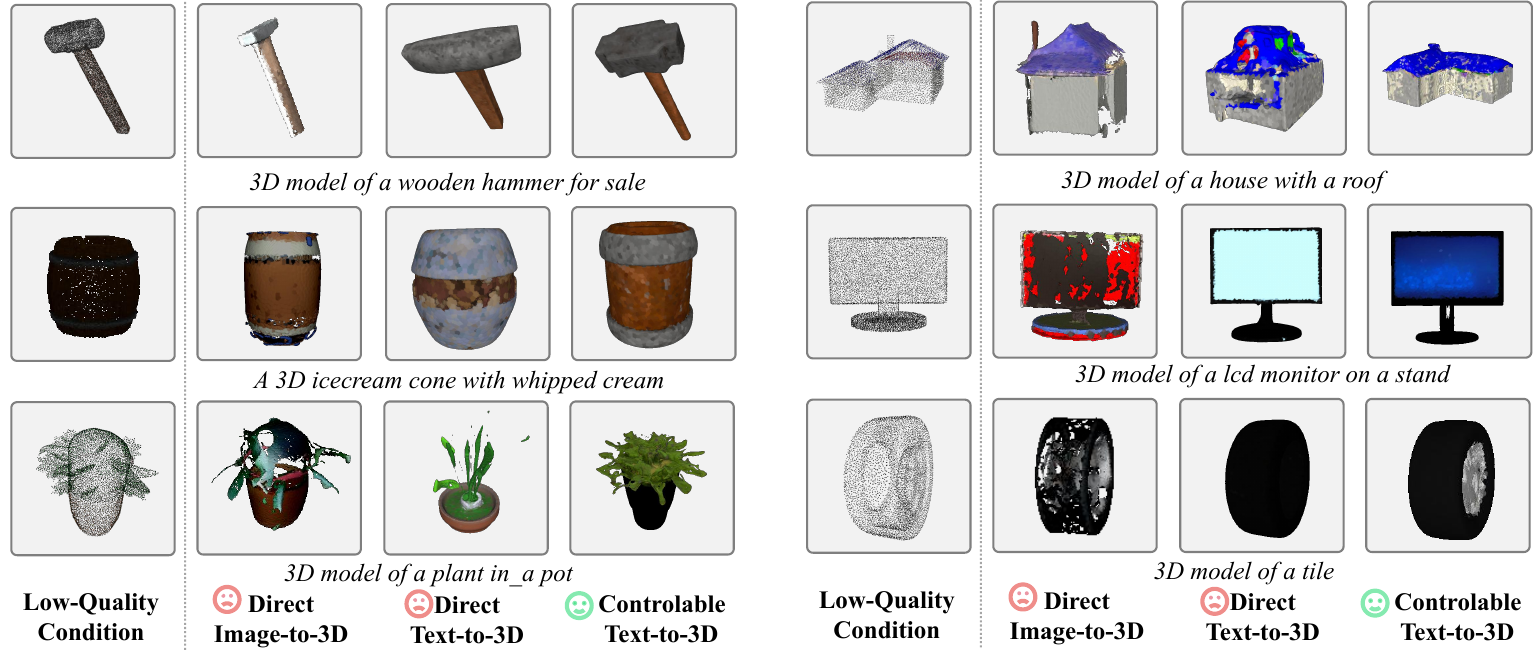}
    \vspace{0.2mm}
    \caption{\textbf{Point Diffusion Results: our controllable text-to-3D.} Given a low-quality point cloud prior, it can generate outcomes superior to direct text-to-3D and image-to-3D methods and more closely align with the low-quality priors, demonstrating controllability.}
    \label{sup_fig5_point_diffusion_results}
    \vspace{-3mm}
\end{figure*}

\section{Additional Experiments}
\label{sup_sec_E Additional Experiments}
~~~In this section, we supplement the details of the experiments. First in Sec.~\ref{sup_sec_E_1 Training Details}, we list all the hyperparameters through the table. Then We give more qualitative results about our experiments. Sec.~\ref{sup_sec_E_2 Point-text Captions and QA Demos} shows the text reference tasks like 3D object point caption and QA and Sec.~\ref{sup_sec_E_3 Point Diffusion Results} shows our point diffusion results.

\subsection{Training Details}
\label{sup_sec_E_1 Training Details}

\vspace{-2mm}
\begin{table}[!h]
    \centering
    \vspace{0pt}

    \begin{tabular}{l|c}
    \toprule
    \multicolumn{1}{c|}{\bfseries Hyperparameters\mdseries} & \textbf{Value/Type} \\
    \midrule
    
    batchsize & 32 \\
    
    training epochs & 10 \\
    
    optimizer & AdamW \\
    
    init lr & 1e-4 \\

    min lr & 1e-5 \\

    warmup lr & 1e-6 \\
    
    weight decay & 0.05 \\
    
    lr schedule & cosine annealing \\
    
    warmup type & linear \\
    
    warmup iters & 5000 \\
    
    \midrule
    
    Point size & 8192 \\
    
    Q-Former queries & 32 \\
    
    \bottomrule
    \end{tabular}

    \vspace{-2mm}
    \caption{\textbf{Training settings and hyperparameters for Stage1.}} 
    \label{sup_tab2_hyperparameters}

\end{table}
\vspace{-3mm}
We detail the hyperparameters of GPT4Point, largely mirroring those used in BLIP-2~\cite{BLIP-2} during the pretrain stage. These parameters are maintained for Stage1: Point-text alignment and the LLM branch in Stage2.~\cref{sup_tab2_hyperparameters} lists them. The parameters for the LLM branch in Stage2 are almost identical to those of Stage1, except for the warm-up iterations, which changed from 5K to 2K. For BLIP-2, after pretraining on multiple datasets, fine-tuning is performed on a smaller dataset and subtasks. Additionally, different image backbones were used in the pretraining and fine-tuning phases. But in our GPT4Point, we only use the pretrain stage in the BLIP-2 and all tasks are evaluated by zero-shot. For the diffusion branch, we need to make the learning rate very small because here we only train the fully connected layers. The init, min and the warmup learing rate is 1e-7, 0 and 1e-8, and we only train 1 epoch.

\subsection{Point-text Captions and QA Demos}
\label{sup_sec_E_2 Point-text Captions and QA Demos}
~~~In this section, we show more point-text qualitative results of GPT4Point. More specific examples are presented in~\cref{sup_fig10_qa_demo}. We can see that GPT4Point is capable of effectively understanding point clouds and can engage in fluent conversations with humans. 

\subsection{Point Diffusion Results}
\label{sup_sec_E_3 Point Diffusion Results}
~~~\cref{sup_fig5_point_diffusion_results} shows more qualitative results of point diffusion results of GPT4Point. We find that GPT4Point can guide text-to-3D processes, generating results with more accurate colors and geometric shapes.

\vspace{-2mm}
\begin{table}[!h]
    \centering
    \vspace{0pt}

    \begin{tabular}{l|c}
    \toprule
    \multicolumn{1}{c|}{\bfseries Content\mdseries} & \textbf{Figure} \\
    \midrule
    Sec.~\ref{sup_sec_3_1 Pyramid-XL: Data Annotation Engine}: Prompts of Level 2 caption & \cref{sup_fig6_prompt1} \\

    Sec.~\ref{sup_sec_3_1 Pyramid-XL: Data Annotation Engine}: Prompts of Level 3 long QA & \cref{sup_fig7_prompt2} \\    
    
    Sec.~\ref{sup_sec_3_1 Pyramid-XL: Data Annotation Engine}: Level 3 caption finetune Point-E & \cref{sup_fig11_dataset_pointe_ft_results} \\
    
    \midrule
    
    Sec.~\ref{sup_sec_D_1 ObjaverseXL-LVIS QA 1K dataset}: ObjaverseXL-LVIS QA 1K & \cref{sup_fig8_qa_dataset} \\
    
    Sec.~\ref{sup_sec_D_2 Anomalous Objects: Generation Failure Cases}: Generation Failure Cases & \cref{sup_fig9_qa_special} \\

    \midrule
    
    Sec.~\ref{sup_sec_E_2 Point-text Captions and QA Demos}: Point-text Captions and QA & \cref{sup_fig10_qa_demo} \\
    
    Sec.~\ref{sup_sec_E_3 Point Diffusion Results}: Point Diffusion Results & \cref{sup_fig5_point_diffusion_results} \\
    
    \bottomrule
    \end{tabular}

    \vspace{-2mm}
    \caption{\textbf{Chapter-Experiment Result Image Correspondence.}} 
    \label{sup_tab3_index_fig}

\end{table}
\vspace{-3mm}

\begin{figure*}[!h]
    \centering
    \includegraphics[width=0.91\linewidth]{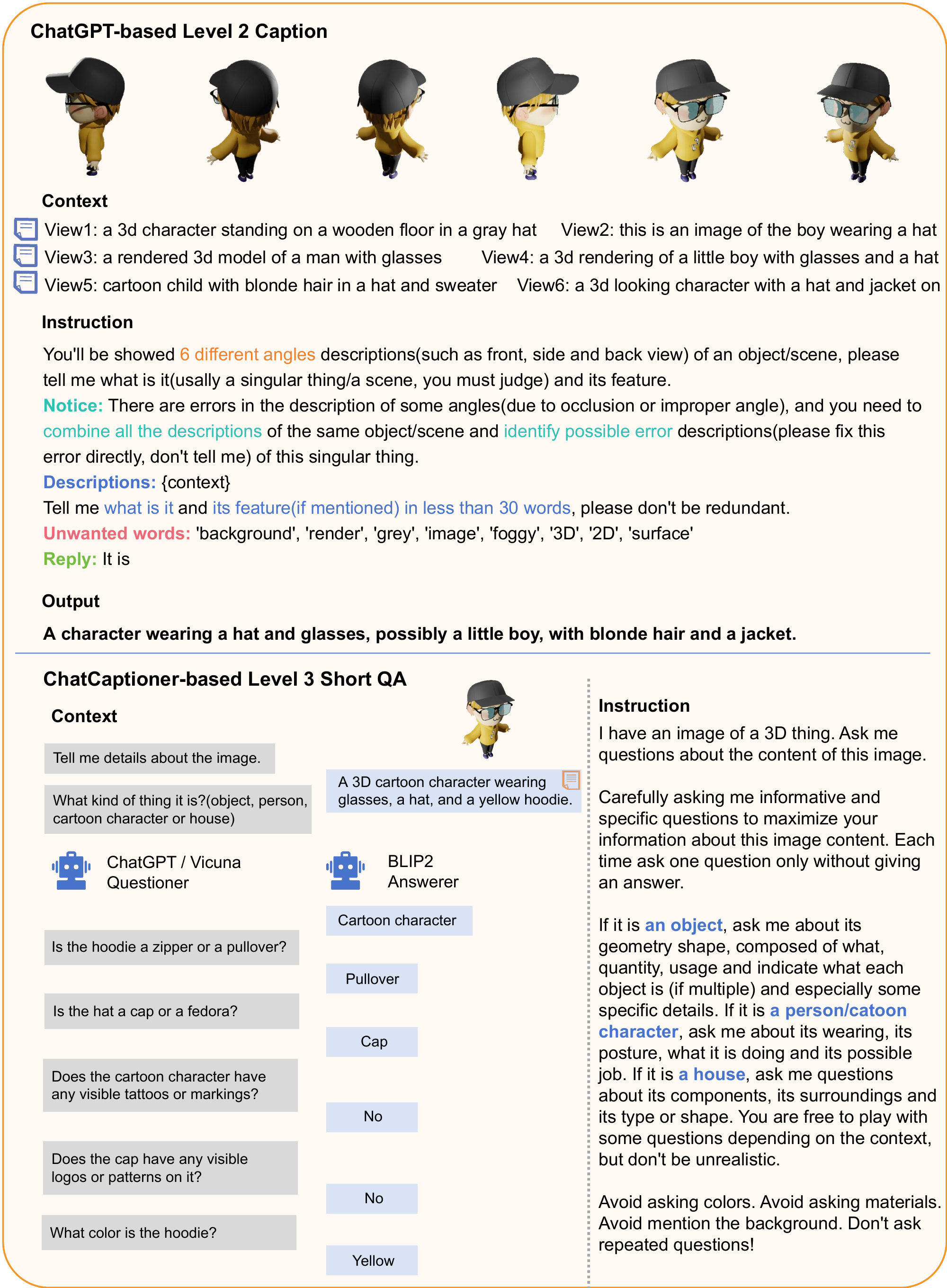}
    \vspace{0.2mm}
    \caption{\textbf{Prompts of Level 2 Caption and Level 3 Short Q\&A in Pyramid-XL.} We present the construction process of GPT-based Level 2 Caption and ChatCaptioner-based Level 3 Short Q\&A, along with the prompts utilized, consisting of context and instruction.}
    \label{sup_fig6_prompt1}
    \vspace{-3mm}
\end{figure*}

\begin{figure*}[!h]
    \centering
    \includegraphics[width=0.91\linewidth]{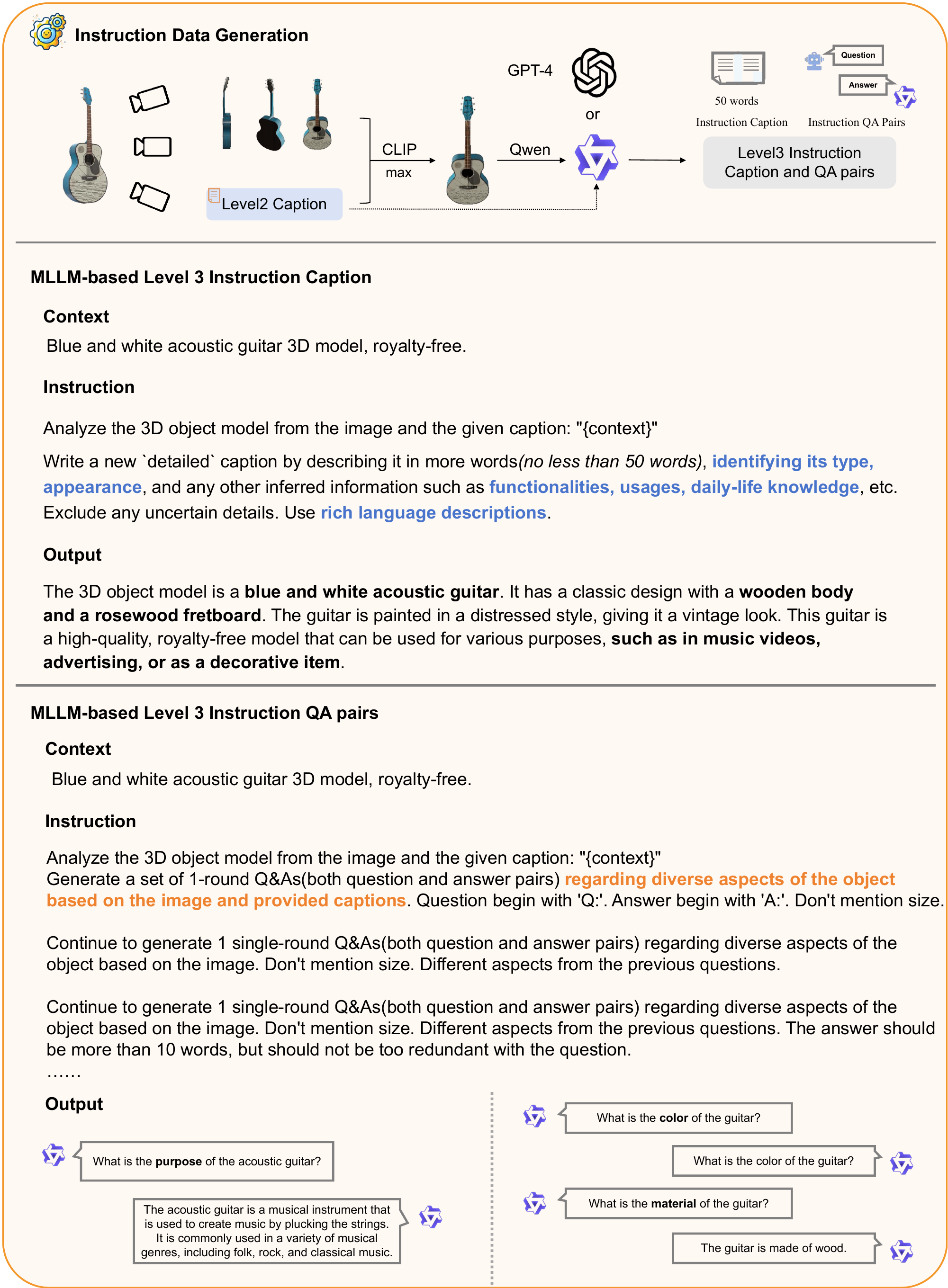}
    \vspace{0.2mm}
    \caption{\textbf{Prompts of MLLM-based Level 3 Instuction Caption and QA pairs in Pyramid-XL.} The top part details the process of constructing the dataset, while below are the specific instructions provided to the MLLM (Qwen-VL[xx]) and the model output.}
    \label{sup_fig7_prompt2}
    \vspace{-3mm}
\end{figure*}

\begin{figure*}[!h]
    \centering
    \includegraphics[width=\linewidth]{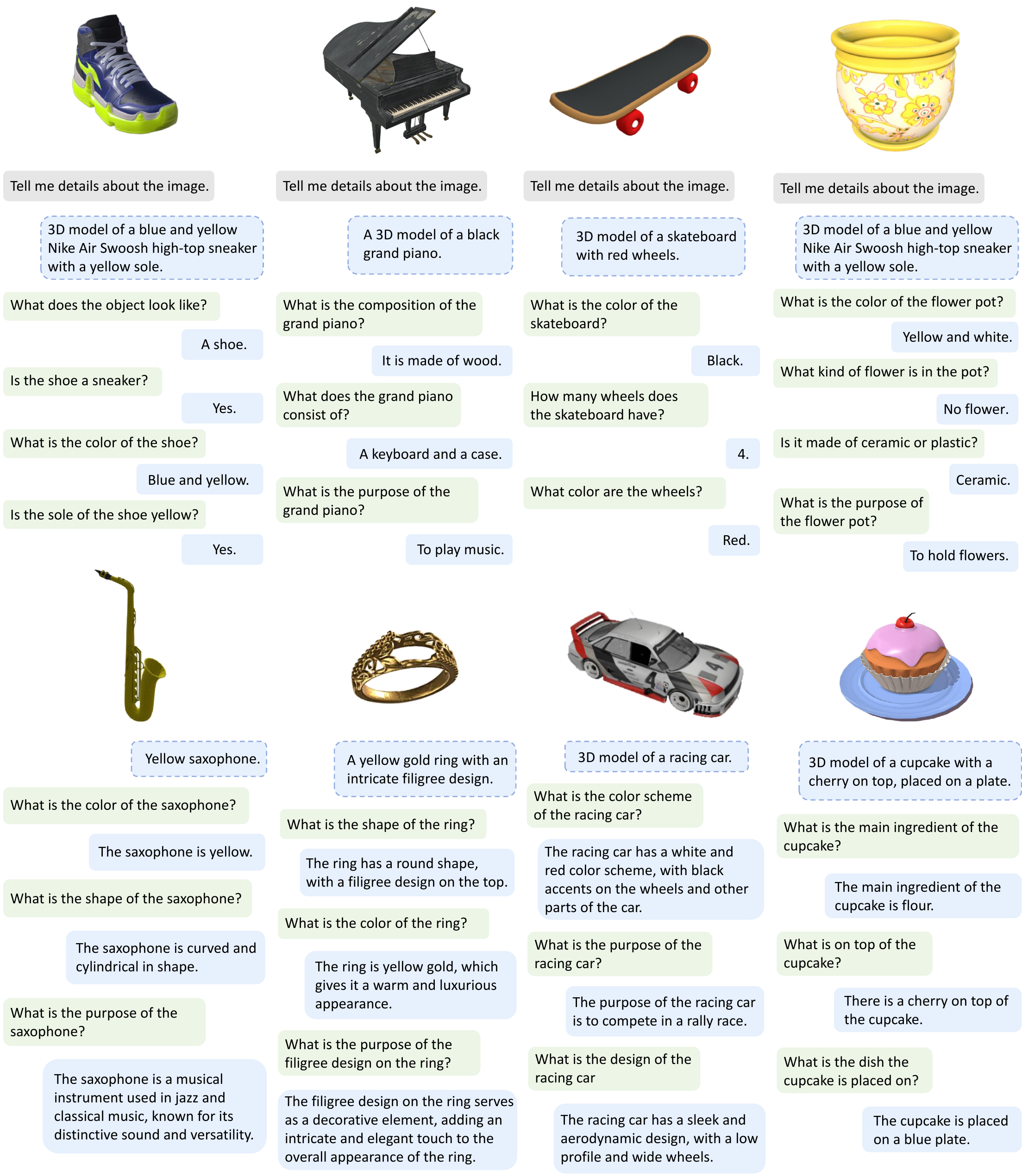}
    \vspace{0.2mm}
    \caption{\textbf{Objaverse-XL QA dataset.} The upper section of the dataset features short Q\&A samples, and the lower part includes long Q\&A samples, covering color, shape, type, material, and purpose. The short Q\&A dataset evaluates performance, while the long Q\&A is for fine-tuning, promoting more detailed language generation and promote the understanding and conversational capabilities.}
    \label{sup_fig8_qa_dataset}
    \vspace{-3mm}
\end{figure*}

\begin{figure*}[!h]
    \centering
    \includegraphics[width=0.92\linewidth]{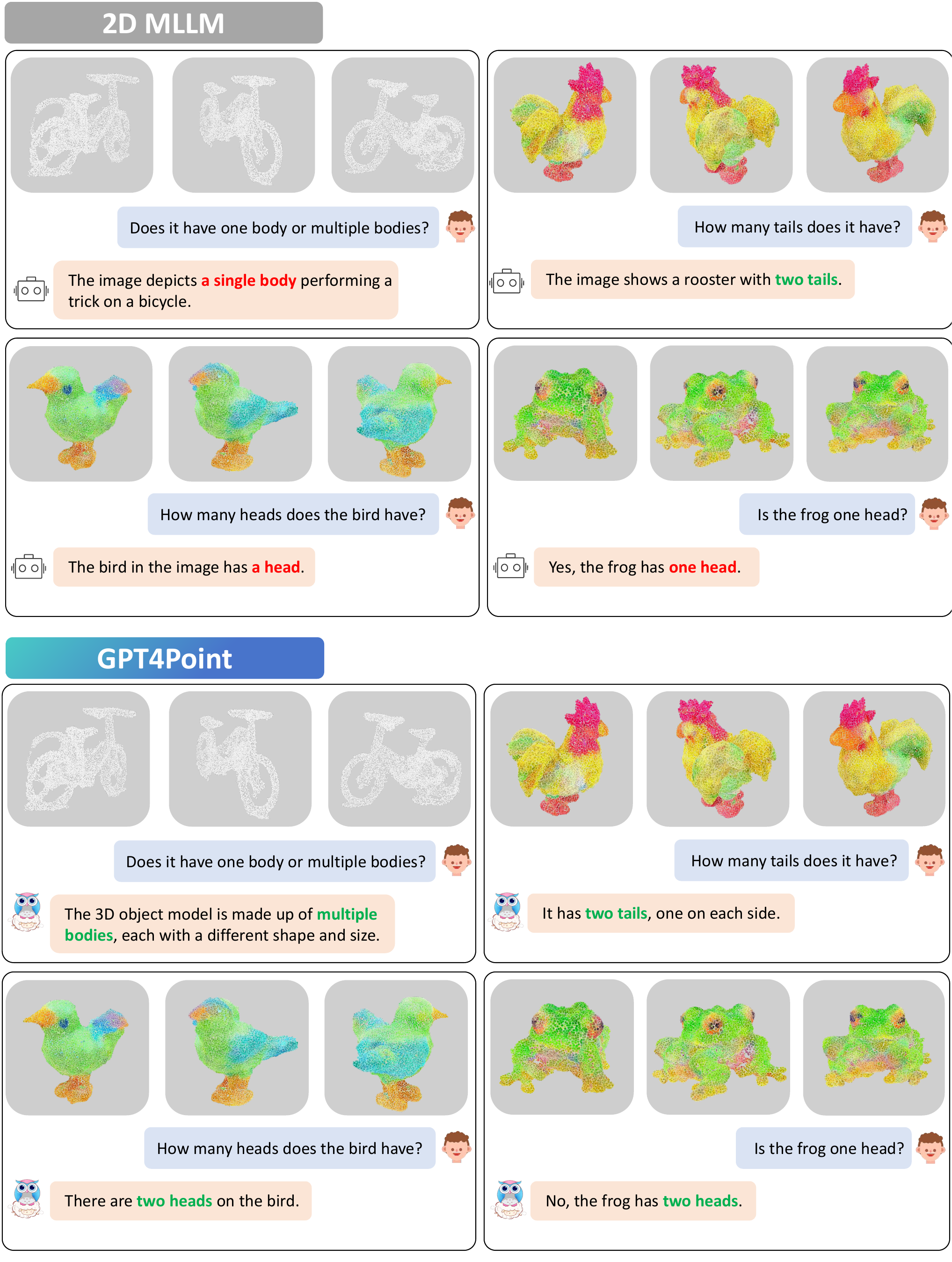}
    \vspace{0.2mm}
    \caption{\textbf{Anomalous Objects: Generation Failure Cases.} The upper and lower parts respectively depict the performance of 2D MLLM and GPT4Point in identifying abnormally generated objects with multi-body and multi-head structures. GPT4Point is effective in making accurate judgments, whereas 2D MLLM, due to the lack of information from single-view images, fails to identify most cases.}
    \label{sup_fig9_qa_special}
    \vspace{-3mm}
\end{figure*}

\begin{figure*}[!h]
    \centering
    \includegraphics[width=0.98\linewidth]{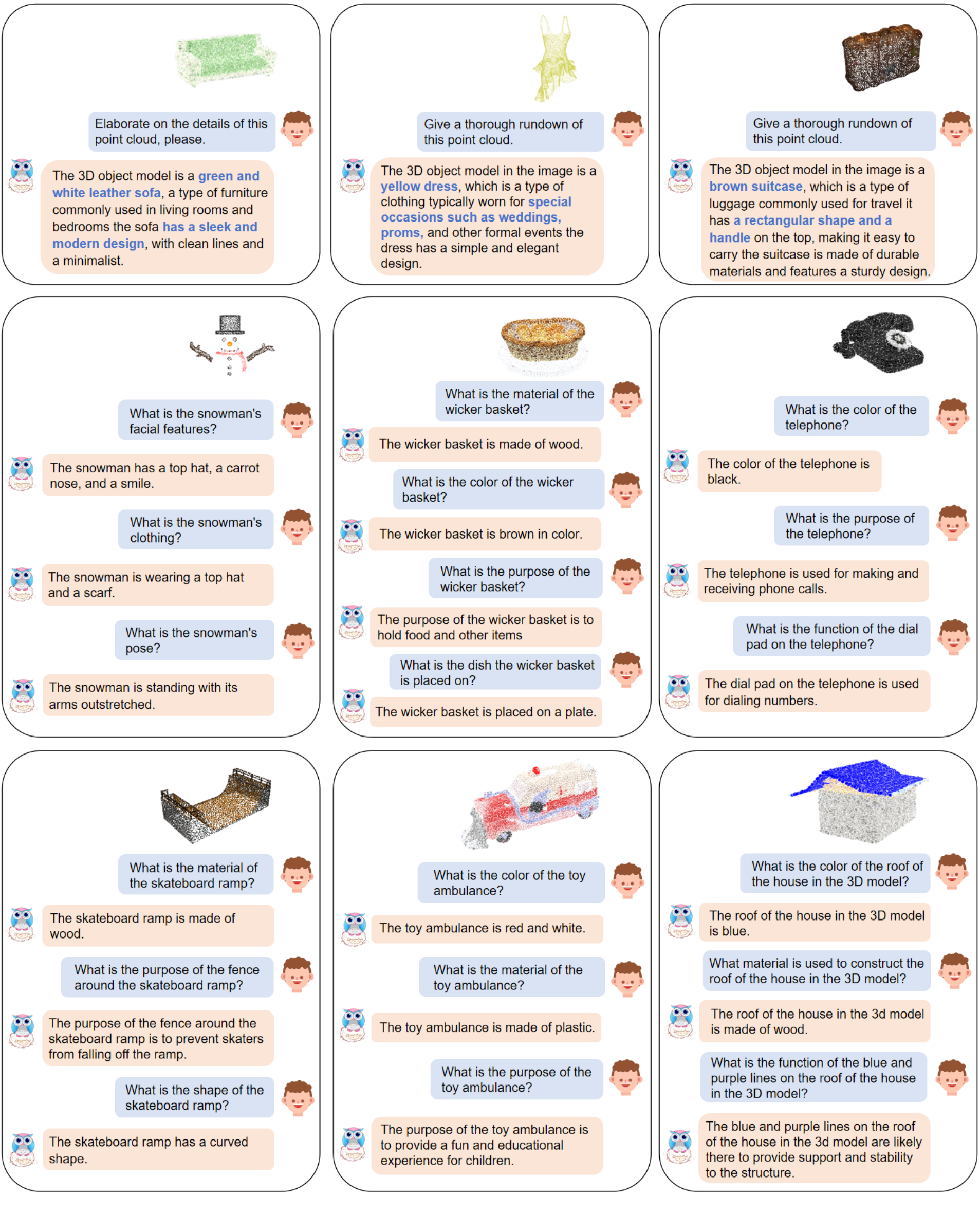}
    \vspace{0.2mm}
    \caption{\textbf{Point-text Captions and QA Demos.} We use the fine-tuned GPT4Point with OPT6.7B model to generate results on the test set, demonstrating that our model performs well on dense captioning tasks and long (instruction) question answering. The results shows our model's capability to comprehend information such as object color and geometry.}
    \label{sup_fig10_qa_demo}
    \vspace{-3mm}
\end{figure*}

\begin{figure*}[!h]
    \centering
    \includegraphics[width=0.77\linewidth]{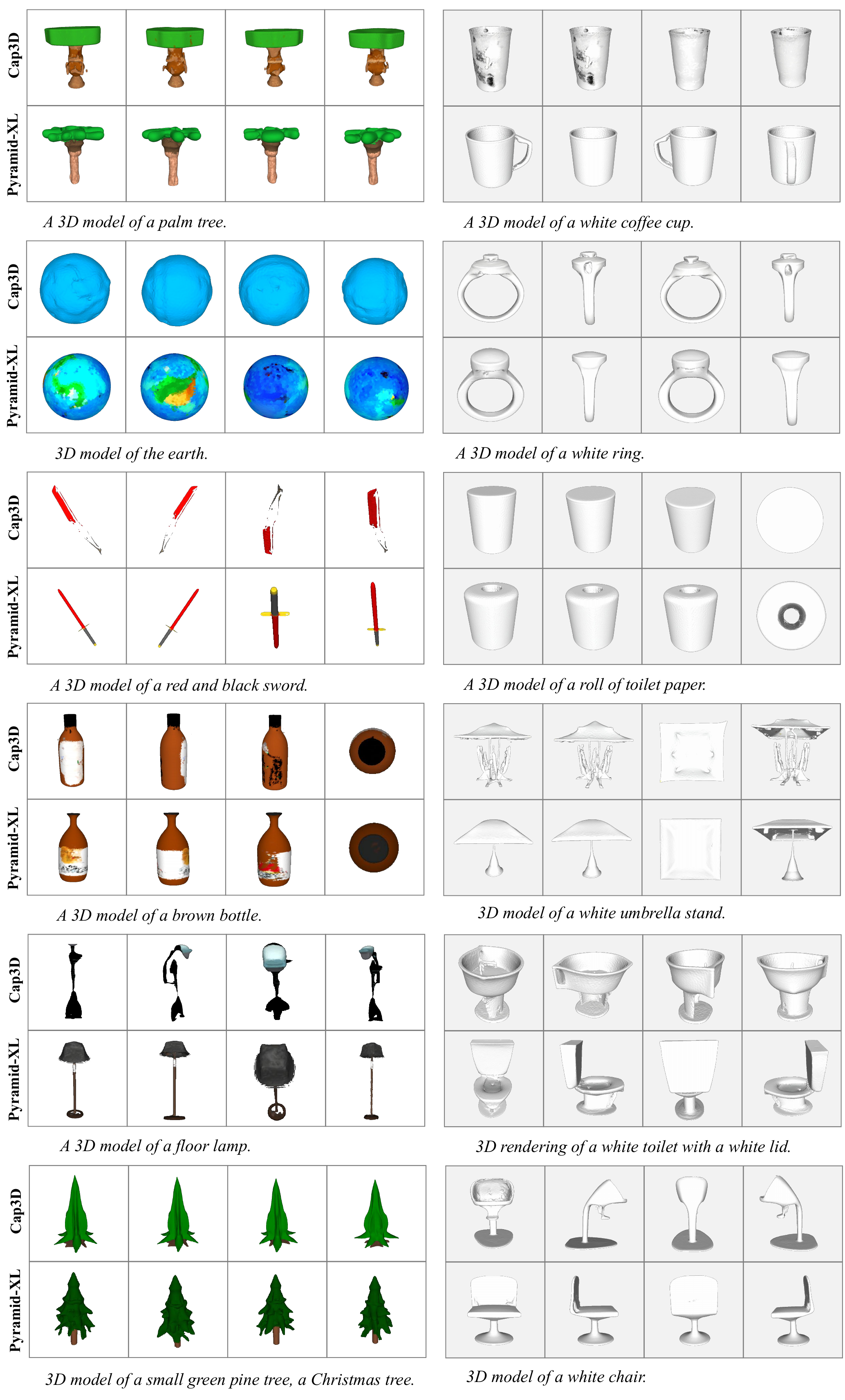}
    \vspace{-4mm}
    \caption{\textbf{Pyramid-XL Level 3 Point-E Finetune Results.} We found that the results of fine-tuning with dense captions from our Pyramid-XL significantly outperform those fine-tuned with Cap3D captions, demonstrating the greater accuracy of our generated captions.}
    \label{sup_fig11_dataset_pointe_ft_results}
    \vspace{-3mm}
\end{figure*}

{
\clearpage
\small
\bibliographystyle{ieee_fullname}

}

\end{document}